\documentclass[10pt,twocolumn,letterpaper]{article}

\usepackage{cvpr}
\usepackage{times}
\usepackage{epsfig}
\usepackage{graphicx}
\usepackage{amsmath}
\usepackage{amssymb}
\usepackage{multirow}
\usepackage{subfigure} 
\usepackage{booktabs}
\usepackage{cite}
\usepackage{xcolor}
\usepackage{caption}

\usepackage[pagebackref=true,breaklinks=true,letterpaper=true,colorlinks,linkcolor=blue,bookmarks=false]{hyperref}

\cvprfinalcopy 

\def\cvprPaperID{6312} 

\ifcvprfinal\fi
\begin{document}


\title{Cascade Cost Volume for High-Resolution Multi-View Stereo \\
and Stereo Matching}

\author{
Xiaodong Gu$^{1}$\footnotemark[1]\hspace{1cm}Zhiwen Fan$^{1}$\footnotemark[1]\hspace{1cm}Zuozhuo Dai$^{1}$\hspace{1cm}Siyu Zhu$^{1}$\hspace{1cm}Feitong Tan$^{12}$\footnotemark[2]\hspace{1cm}Ping Tan$^{12}$\\
${}^{1}$Alibaba A.I. Labs\hspace{1.5cm}${}^{2}$Simon Fraser University
\vspace{-20mm}
}



\twocolumn[{%
\maketitle
\renewcommand\twocolumn[1][]{#1}%
\begin{center}
    \includegraphics[width=\textwidth]{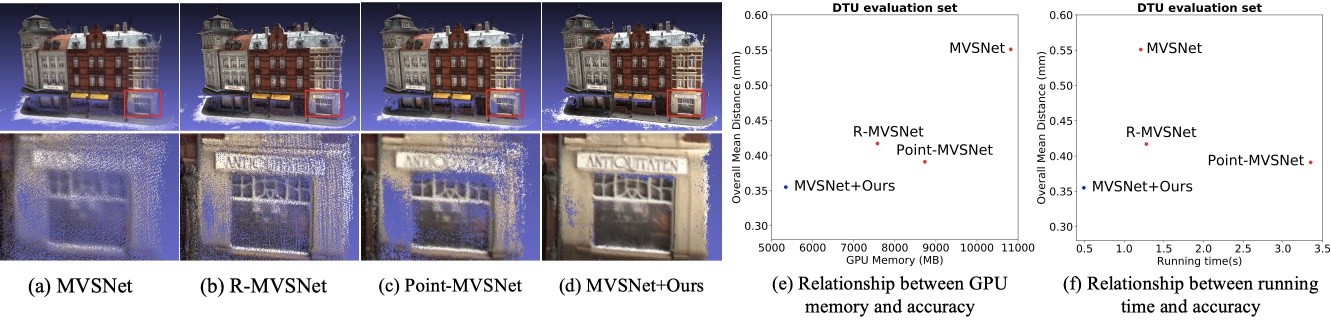}\vspace{-3mm}
    \captionof{figure}{Comparison between the state-of-the-art learning-based multi-view stereo approaches~\cite{chen2019point,yao2018mvsnet,yao2019recurrent} and MVSNet+Ours. (a)-(d): Reconstructed point clouds of MVSNet~\cite{yao2018mvsnet}, R-MVSNet~\cite{yao2019recurrent}, Point-MVSNet~\cite{chen2019point} and MVSNet+Ours. (e) and (f): The relationship between reconstruction accuracy and GPU memory or run-time. The resolution of input images is 1152 $\times$ 864.}
    \vspace{-3mm}
    \label{fig:teaser}
\end{center}%

}]

\renewcommand{\thefootnote}{\fnsymbol{footnote}} 
\footnotetext[1]{Equal contribution.} 
\footnotetext[2]{This work was done during an internship at Alibaba A.I. Labs.} 

\begin{abstract}\vspace{-4mm}

    The deep multi-view stereo (MVS) and stereo matching approaches generally construct 3D cost volumes to regularize and regress the depth or disparity.
    These methods are limited with high-resolution outputs since the memory and time costs grow cubically as the volume resolution increases.
    In this paper, we propose a memory and time efficient cost volume formulation complementary to existing multi-view stereo and stereo matching approaches based on 3D cost volumes.
    First, the proposed cost volume is built upon a feature pyramid encoding geometry and context at gradually finer scales.
    Then, we can narrow the depth (or disparity) range of each stage by the prediction from the previous stage.
    With gradually higher cost volume resolution and adaptive adjustment of depth (or disparity) intervals, the output is recovered in a coarser to fine manner.
    
    
    We apply the cascade cost volume to the representative MVS-Net, and obtain a $35.6\%$ improvement on DTU benchmark (1st place), with $50.6\%$ and $59.3\%$ reduction in GPU memory and run-time.
    It is also rank first on Tanks and Temples benchmark of all deep models.
    The statistics of accuracy, run-time and GPU memory on other representative stereo CNNs also validate the effectiveness of our proposed method. Our source code is available at \url{https://github.com/alibaba/cascade-stereo}.
\end{abstract}


\section{Introduction}
Convolutional neural networks (CNNs) have been widely adopted in 3D reconstruction and broader computer vision tasks.
State-of-the-art multi-view stereo~\cite{im2019dpsnet, yao2018mvsnet, yao2019recurrent, Luo_2019_ICCV} and stereo matching algorithms~\cite{kendall2017end, chang2018pyramid, guo2019group, zhang2019ga, wu2019iccv_semanticstereo, nie2019multi} often compute a 3D cost volume according to a set of hypothesized depth (or disparity) and warped features. 3D convolutions are applied to this cost volume to regularize and regress the final scene depth (or disparity).

Compared with the methods based on 2D CNNs~\cite{mayer2016large, zbontar2016stereo}, the 3D cost volume can capture better geometry structures, perform photometric matching in 3D space, and alleviate the influence of image distortion caused by perspective transformation and occlusions \cite{chen2019point}.
However, methods relying on 3D cost volumes are often limited to low-resolution input images (and results),
because 3D CNNs are generally time and GPU memory consuming.
Typically, these methods downsample the feature maps to formulate the cost volumes at a lower resolution~\cite{kendall2017end, chang2018pyramid, guo2019group, zhang2019ga, wu2019iccv_semanticstereo, nie2019multi, im2019dpsnet, yao2018mvsnet, yao2019recurrent, Luo_2019_ICCV, chen2019point}, and adopt upsampling~\cite{kendall2017end, chang2018pyramid, guo2019group, zhang2019ga,nie2019multi, wu2019iccv_semanticstereo,yang2018segstereo, song2018edgestereo} or post-refinement~\cite{Luo_2019_ICCV, chen2019point} to output the final high-resolution result.

Inspired by the previous coarse-to-fine learning-based stereo approaches~\cite{wang2019anytime,tonioni2019real,yin2019hierarchical}, we present a novel cascade formulation of 3D cost volumes.
We start from a feature pyramid to extract multi-scale features which are commonly used in standard multi-view stereo~\cite{yao2018mvsnet} and stereo matching~\cite{guo2019group, chang2018pyramid} networks.
In a coarse-to-fine manner, the cost volume at the early stages is built upon larger scale semantic 2D features with sparsely sampled depth hypotheses, which lead to a relatively lower volume resolution.
Subsequently, the later stages use the estimated depth (or disparity) maps from the earlier stages to adaptively adjust the sampling range of depth (or disparity) hypotheses and construct new cost volumes where finer semantic features are applied.
This adaptive depth sampling and adjustment of feature resolution ensures the computation and memory resources are spent on more meaningful regions. In this way, our cascade structure can remarkably decrease computation time and GPU memory consumption.
The effectiveness of our method can be seen in Figure~\ref{fig:teaser}.



We validate our method on both multi-view stereo and stereo matching on various benchmark datasets. For multi-view stereo, our cascade structure achieves the best performance on the DTU dataset~\cite{aanaes2016dtu} at the submission time of this paper, when combined with  MVSNet~\cite{yao2018mvsnet}.
It is also the state-of-the-art
learning-based method on Tanks and Temples benchmark~\cite{knapitsch2017tanks}.
For stereo matching, our method reduces the end-point-error (EPE) and GPU memory consumption of GwcNet~\cite{guo2019group} by about 15.2\% and 36.9\% respectively.

\begin{figure*}[ht]
\begin{center}
\includegraphics[width=\linewidth]{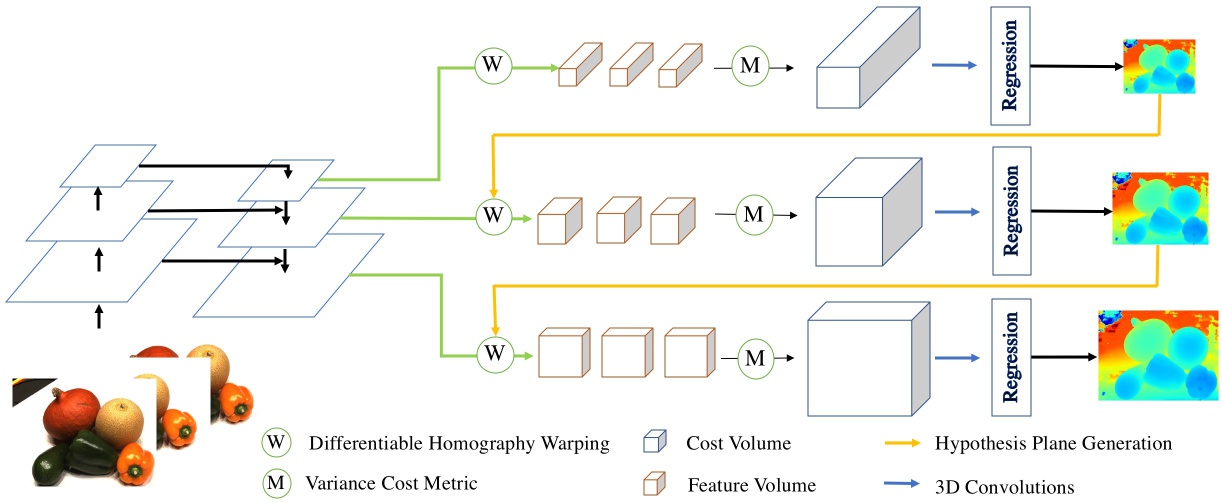}
\end{center}
\vspace{-4mm}
   \caption{Network architecture of the proposed cascade cost volume on MVSNet~\cite{yao2018mvsnet}, denoted as MVSNet+Ours.}
\vspace{-4mm}
\label{fig:arc}
\end{figure*}

\section{Related Work} 
\paragraph{Stereo Matching}
According to the survey by Scharstein $et$ $al.$\cite{scharstein2002taxonomy}, a typical stereo matching algorithm contains four steps: matching cost calculation, matching cost aggregation, disparity calculation, and disparity refinement. Local methods~\cite{zhang2009cross, yang2012non, mei2013segment_gwc20} aggregate matching costs with neighboring pixels and usually utilize the winner-take-all strategy to choose the optimal disparity. Global methods \cite{sun2003stereo_gwc30,klaus2006segment_gwc13,hirschmuller2005accurate_gwc7} construct an energy function and try to minimize it to find the optimal disparity.
More specifically, works in \cite{sun2003stereo_gwc30,klaus2006segment_gwc13} use  belief propagation and semi-global matching \cite{hirschmuller2005accurate_gwc7} to approximate the global optimization with dynamic programming. 

In the context of deep neural networks, CNNs based stereo matching methods are first introduced by Zbontar and LeCun \cite{zbontar2015computing}, in which a convolutional neural network is introduced to learn the similarity measure of small patch pairs. The introduction of the widely used 3D cost volume in stereo is first proposed in GCNet \cite{kendall2017end},
in which the disparity regression step uses the soft argmin operation to figure out the best matching results. 
PSMNet \cite{chang2018pyramid} further introduces pyramid spatial pooling and 3D hourglass networks for cost volume regularization and yields better results. GwcNet \cite{guo2019group} modifies the structure of 3D hourglass and introduces group wise correlation to form a group based 3D cost volume. 
HSM~\cite{yang2019hierarchical} builds a light model for high-resolution images with a hierarchical design.
EMCUA~\cite{nie2019multi} introduces an approach for multi-level context ultra-aggregation.
GANet~\cite{zhang2019ga} constructs several semi-global aggregation layers and local guided aggregation layers to further improve the accuracy. DeepPruner~\cite{duggal2019deeppruner} is a coarse to fine method which proposes a differentiable PatchMatch-based module to predict the pruned search range for each pixel.

Although methods based on 3D cost-volume remarkably boost the performance, they are limited to downsampled cost volumes and rely on interpolation operations to generate high-resolution disparity. Our cascade cost volumes can be combined with these methods to improve the disparity accuracy and GPU memory efficiency.


\paragraph{Multi-View Stereo}\vspace{-4mm}
According to the comprehensive survey~\cite{furukawa2015multi}, works in traditional muti-view stereo can be roughly categorised into volumetric methods ~\cite{kutulakos2000theory_mvsnet20,seitz1999photorealistic_mvsnet33,ji2017surfacenet_mvsnet14,kar2017learning_mvsnet15}, which estimate the relationship between each voxel and surfaces; point cloud based methods~\cite{lhuillier2005quasi_22,furukawa2009accurate_7}, which directly process 3D points to iteratively densify the results; and depth map reconstruction methods~ \cite{tola2012efficient_mvsnet35,campbell2008using_mvsnet3,galliani2015massively_mvsnet8,schonberger2016pixelwise_mvsnet32,yao2017relative_mvsnet38, romanoni2019tapa}, which use only one reference and a few source images for single depth map estimation. For large-scale Structure-from-Motion, works in~\cite{zhang2017distributed, zhu2018very} use distributed methods based on distributed motion averaging and global camera consensus.

Recently, learning-based approaches also demonstrate superior performance on multi-view stereo. Multi-patch similarity~\cite{hartmann2017learned} introduces a learned cost metric. SurfaceNet~\cite{ji2017surfacenet_mvsnet14} and DeepMVS~\cite{huang2018deepmvs} pre-warp the multi-view images to 3D space and use deep networks for regularization and aggregation. Most recently, multi-view stereo based on 3D cost volumes have been proposed in~\cite{yao2018mvsnet, im2019dpsnet, Luo_2019_ICCV, yao2019recurrent, chen2019point, hou2019multi, xue2019mvscrf}. A 3D cost volume is built based on warped 2D image features from multiple views and 3D CNNs are applied for cost regularization and depth regression.
Because the 3D CNNs require large GPU memory, these methods generally use downsampled cost volumes. 
Our cascade cost volume can be easily integrated into these methods to enable high-resolution cost volumes and further boosts accuracy, computational speed, and GPU memory efficiency.

\paragraph{High-Resolution Output in Stereo and MVS}\vspace{-4mm}
Recently, some learning-based methods try to reduce the memory requirement in order to generate high resolution outputs.
Instead of using voxel grids, Point MVSNet~\cite{chen2019point} proposes to use a small cost volume to generate the coarse depth and uses a point-based iterative refinement network to output the full resolution depth.
In comparison, a standard MVSNet combined with our cascade cost volume can output full resolution depth with superior accuracy using less run-time and GPU memory than Point MVSNet~\cite{chen2019point}.
Works in~\cite{wang2017ocnn, riegler2017octnet} partition advanced space to reduce memory consumption and construct a fixed cost volume representation which lacks flexibility.
Works in~\cite{yang2018segstereo, song2018edgestereo, Luo_2019_ICCV} build extra refinement module by 2D CNNs and output a high resolution prediction. 
Notably, such refinement modules can be utilized jointly with our proposed cascade cost volume. 
 

\section{Methodology} 
This section describes the detailed architecture of the proposed cascade cost volume which is complementary to the existing 3D cost volume based methods in multi-view stereo and stereo matching.
Here, we use the representative MVSNet~\cite{yao2018mvsnet} and PSMNet~\cite{chang2018pyramid} as the backbone networks to demonstrate the application of the cascade cost volume in multi-view stereo and stereo matching tasks respectively.
Figure~\ref{fig:arc} shows the architecture of MVSNet+Ours.

\subsection{Cost volume Formulation}
Learning-based multi-view stereo~\cite{yao2018mvsnet,yao2019recurrent,chen2019point} and stereo matching \cite{chang2018pyramid,kendall2017end,zbontar2015computing,zhang2019ga,guo2019group} construct 3D cost volumes to measure the similarity between 
corresponding image patches and determine whether they are matched.
Constructing 3D cost volume requires three major steps in both multi-view stereo and stereo matching.
First, the discrete hypothesis depth (or disparity) planes are determined.
Then, we warp the extracted 2D features of each view to the hypothesis planes and construct the feature volumes, which are finally fused together to build the 3D cost volume.
Pixel-wise cost calculation is generally ambiguous in inherently ill-posed regions such as occlusion areas, repeated patterns, textureless regions, and reflective surfaces.
To solve this, 3D CNNs at multiple scales are generally introduced to aggregate contextual information and regularize the possibly noise-contaminated cost volumes.

\vspace{-4mm}
\paragraph{3D Cost Volumes in Multi-View Stereo}
MVSNet \cite{yao2018mvsnet} proposes to use fronto-parallel planes at different depth as hypothesis planes and the depth range is generally determined by the sparse reconstruction.
The coordinate mapping is determined by the homography:
\begin{equation}
H_{i}(d)=K_{i} \cdot R_{i}\cdot (I - \frac{(t_{1}-t_{i})\cdot{n_{1}}^{T}}{d}) \cdot {R_{1}}^{T} \cdot {K_{1}}^{-1}
\label{eq_mvsmapping}
\end{equation}
where $H_{i}(d)$ refers to the homography between the feature maps of the $i^{th}$ view and the reference feature maps at depth $d$.
Moreover, $K_{i},R_{i}, t_{i}$ refers to the camera intrinsics, rotations and translations of the $i^{th}$ view respectively, and $n_{1}$ denotes the principle axis of the reference camera.
Then differentiable homography is used to warp 2D feature maps into hypothesis planes of the reference camera to form feature volumes.
To aggregate multiple feature volumes to one cost volume, the variance-based cost metric is proposed to adapt an arbitrary number of input feature volumes.

\paragraph{3D Cost Volumes in Stereo Matching}\vspace{-3mm}
PSMNet \cite{chang2018pyramid} uses disparity levels as hypothesis planes and the range of disparity is designed according to specific scenes.
Since the left and right images have been rectified, the coordinate mapping is determined by the offset in the x-axis direction:
\begin{equation}
C_{r}(d)=X_{l} - d
\label{eq_stereomapping}
\end{equation}
where $C_{r}(d)$ refers to the transformed x-axis coordinate of the right view at disparity $d$,
and $X_{l}$ is the source x-axis coordinate of the left view. To build feature volumes, we warp the feature maps of the right view to the left view using the translation along the x-axis.
There are multiple ways to build the final cost volume.
GCNet~\cite{kendall2017end} and PSMNet~\cite{chang2018pyramid} concatenate the left feature volume and the right feature volume without decreasing the feature dimension.
The work~\cite{zbontar2016stereo} uses the sum of absolute differences to compute matching cost. DispNetC \cite{mayer2016large} computes full correlation about the left feature volume and right feature volume and produces only a single-channel correlation map for each disparity level. 
GwcNet \cite{guo2019group} proposes group-wise correlation by splitting the features into groups and computing correlation maps in each group.

\begin{figure}[]
\begin{minipage}[t]{.3\linewidth}
\centering
\includegraphics[width=0.9\linewidth]{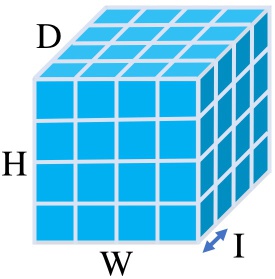}
\end{minipage}
\hfill
\begin{minipage}[t]{.67\linewidth}
\vspace{-18mm}
\resizebox{\linewidth}{!}{%
\begin{tabular}{c|c|c|c}
\midrule[1pt]
&Plane Num.  &Plane Interv.  &Spatial Res.
\\  \midrule[1pt]
\multirow{1}*{Efficiency}     & Negative        &  Positive             &Negative    \\ \midrule[0.5pt]
\multirow{1}*{Accuracy}     & Positive        &  Negative         &Positive    \\ 
\midrule[1pt]
\end{tabular}
}
\end{minipage}
\vspace{-3mm}
\caption{\textbf{Left}: the standard cost volume. D is the number of hypothesis planes, W $\times$ H is the spatial resolution and I is the plane interval. \textbf{Right}: The influence factors of efficiency (run-time and GPU memory) and accuracy.}
\vspace{-6mm}
\label{fig:efficiency_accuracy}
\end{figure}

\subsection{Cascade Cost Volume}
Figure~\ref{fig:efficiency_accuracy} shows a standard cost volume of a resolution of $W \times H \times  D \times F$, where $W\times H$ denotes the spatial resolution, $D$ is the number of plane hypothesis, and $F$ is the channel number of feature maps.
As mentioned in~\cite{yao2018mvsnet, yao2019recurrent, chen2019point}, 
an increased number of plane hypothesis $D$, a larger spatial resolution $W\times H$, and a finer plane interval are likely to improve the reconstruction accuracy.
However, the GPU memory and run-time grow cubically as the resolution of the cost volume increases. 
As demonstrated in R-MVSNet~\cite{yao2019recurrent}, MVSNet~\cite{yao2018mvsnet} is able to process a maximum cost volume of $H \times W \times D \times F = 1600 \times 1184 \times 256 \times 32$ on a 16 GB Tesla P100 GPU.
To resolve the problems above, we propose a cascade cost volume formulation and predict the output in a coarse-to-fine manner.

\begin{figure}
    \centering
    \vspace{-5mm}
    \includegraphics[width=0.8\linewidth]{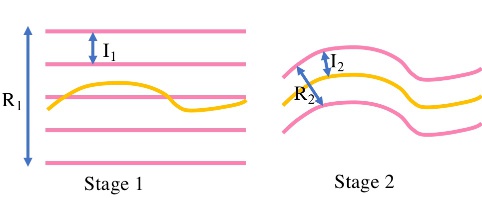}
    \vspace{-3mm}
    \caption{Illustration of hypothesis plane generation. $R_k$ and $I_k$ are respectively the hypothesis range and the hypothesis plane number at the $k^{th}$ stage. Pink lines are hypothesis planes. Yellow line indicates the predicted depth (or disparity) map from stage 1, which is used to determine the hypothesis range and hypothesis plane intervals at stage 2.}
    \vspace{-5mm}
    \label{fig:range_interval}
\end{figure}

\paragraph{Hypothesis Range}\vspace{-3mm}
As shown in Figure~\ref{fig:range_interval}, 
the depth (or disparity) range of the first stage denoted by $R_1$ covers the entire depth (or disparity) range of the input scene.
In the following stages, we can base on the predicted output from the previous stage, and narrow the hypothesis range.
Consequently, we have $R_{k+1} = R_{k}\cdot w_{k}$, where $R_k$ is the hypothesis range at the $k^{th}$ stage and $w_{k}<1$ is the reducing factor of hypothesis range.  

\paragraph{Hypothesis Plane Interval}\vspace{-3mm}
We also denote the depth (or disparity) interval at the first stage as $I_1$.
Compared with the commonly adopted single cost volume formulation~\cite{chang2018pyramid, yao2018mvsnet}, the initial hypothesis plane interval is comparatively larger to generate a coarse depth (or disparity) estimation. 
In the following stages, finer hypothesis plane intervals are applied to recover more detailed outputs.
Therefore, we have: $I_{k+1} = I_{k} \cdot p_k$, where $I_k$ is the hypothesis plane interval at the $k^{th}$ stage and $p_k<1$ is the reducing factor of hypothesis plane interval. 

\paragraph{Number of Hypothesis Planes}\vspace{-3mm}
At the $k^{th}$ stage, given the hypothesis range $R_k$ and hypothesis plane interval $I_k$, the corresponding number of hypothesis planes $D_k$ is determined by the equation: $D_k=R_k/I_k$.
When the spatial resolution of a cost volume is fixed, a larger $D_k$ generates more hypothesis planes and correspondingly more accurate results while leads to increased GPU memory and run-time.
Based on the cascade formulation, we can effectively reduce the total number of hypothesis planes since the hypothesis range is remarkably reduced stage by stage while still covering the entire output range. 



\paragraph{Spatial Resolution}\vspace{-3mm}
Following the practices of Feature Pyramid Network~\cite{lin2017feature}, we double the spatial resolution of the cost volume at every stage along with the doubled resolution of the input feature maps. We define N as the total stage number of cascade cost volume, then the spatial resolution of cost volume at the $k^{th}$ stage is defined as $\frac{W}{2^{N-k}}\times \frac{H}{2^{N-k}}$. We set $N=3$ in multi-view stereo tasks and $N=2$ in stereo matching tasks.  

\paragraph{Warping Operation}\vspace{-3mm}
Applying the cascade cost volume formulation to multi-view stereo, we base on Equation~\ref{eq_mvsmapping} and rewrite the homography warping function at the ${(k+1)}^{th}$ stage as:
\vspace{-6mm}
\begin{equation}
H_{i}(d_{k}^{m}+\Delta_{k+1}^{m})=K_{i} \cdot R_{i} \cdot (I - \frac{(t_{1}-t_{i})\cdot{n_{1}}^{T}}{d_{k}^{m}+\Delta_{k+1}^{m}}) \cdot {R_{1}}^{T} \cdot {K_{1}}^{-1}
\label{eq_mvs_casmapping}
\end{equation}
where $d^m_k$ denotes the predicted depth of the $m^{th}$ pixel at the $k^{th}$ stage, and $\Delta_{k+1}^{m}$ is the residual depth of the $m^{th}$ pixel to be learned at the ${k+1}^{th}$ stage.

Similarly in stereo matching, we reformulate Equation~\ref{eq_stereomapping} based on our cascade cost volume.
The $m^{th}$ pixel coordinate mapping at the ${k+1}^{th}$ stage is expressed as: 
\begin{equation}
C_{r}({d_k^{m}}+\Delta_{k+1}^{m})=X_{l} - ({d_k^{m}}+\Delta_{k+1}^{m})
\label{eq_stereo_casmapping}
\end{equation}
where $d^m_k$ denotes the predicted disparity of the $m^{th}$ pixel at the $k^{th}$ stage,
and $\Delta_{k+1}^{m}$ denotes the residual disparity of the $m^{th}$ pixel to be learned at the ${k+1}^{th}$ stage.

\begin{figure*}[!th]
\begin{center}
\def\col{0.195}
{\includegraphics[width=\col\linewidth]{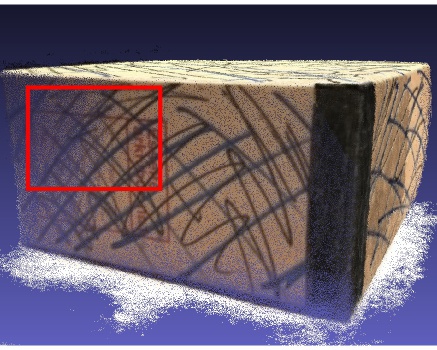}}
{\includegraphics[width=\col\linewidth]{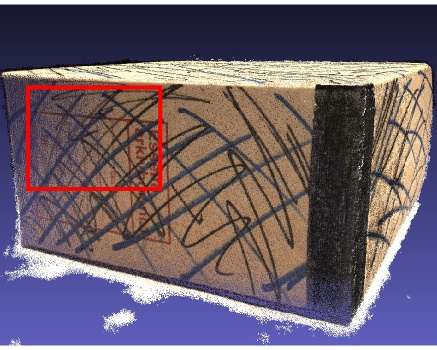}}
{\includegraphics[width=\col\linewidth]{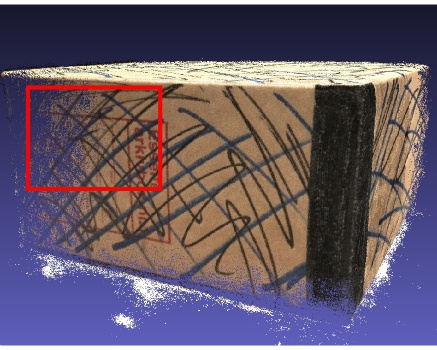}}
{\includegraphics[width=\col\linewidth]{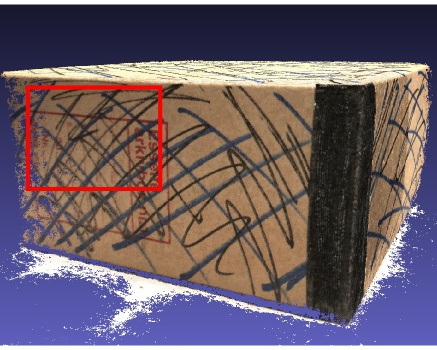}} 
{\includegraphics[width=\col\linewidth]{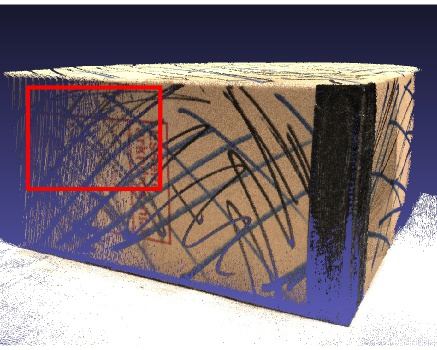}} 
 \subfigure[MVSNet~\cite{yao2018mvsnet}]    
 {\includegraphics[width=\col\linewidth]{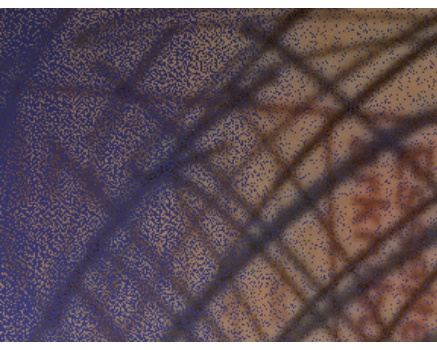}}
 \subfigure[R-MVSNet~\cite{yao2019recurrent}]    {\includegraphics[width=\col\linewidth]{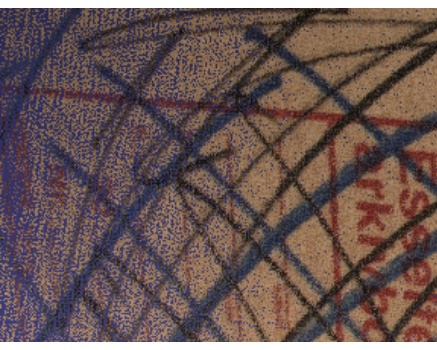}}
 \subfigure[Point MVSNet~\cite{chen2019point}] {\includegraphics[width=\col\linewidth]{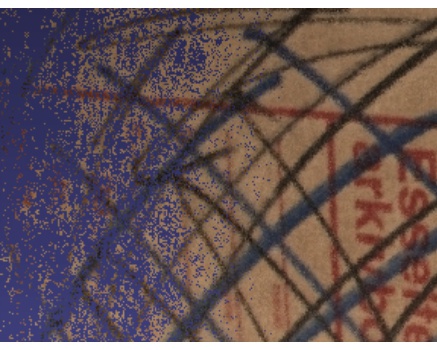}}
 \subfigure[MVSNet+Ours]
 {\includegraphics[width=\col\linewidth]{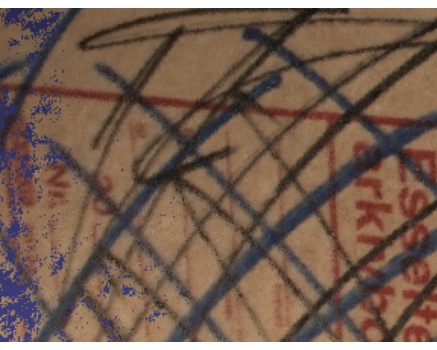}} 
 \subfigure[Ground Truth] 
 {\includegraphics[width=\col\linewidth]{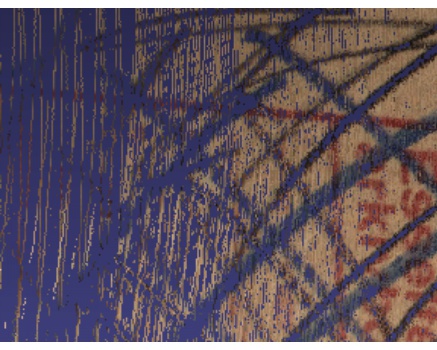}} 
\end{center}
\vspace{-8mm}
 \caption{Multi-view stereo qualitative results of scan 10 on DTU dataset~\cite{aanaes2016dtu}. \textbf{Top row}: Generated point clouds of different methods and ground truth point clouds. \textbf{Bottom row}: Zoomed local areas.}
 \vspace{-2mm}
\label{fig:fig_mvs}
\end{figure*}

\subsection{Feature Pyramid}  

In order to obtain high-resolution depth (or disparity) maps, previous works~\cite{zhang2019ga, wu2019iccv_semanticstereo, nie2019multi, Luo_2019_ICCV} generally generate a comparatively low-resolution depth (or disparity) map using the standard cost volume and then upsample and refine it with 2D CNNs.
The standard cost volume is constructed using the top level feature maps which contains high-level semantic features but lacks low-level finer representations.
Here, we refer to Feature Pyramid Network~\cite{lin2017feature} and adopt its feature maps with increased spatial resolutions to build the cost volumes of higher resolutions.
For example, when applying cascade cost volume to MVSNet~\cite{yao2018mvsnet}, we build three cost volumes from the feature maps \{P1, P2, P3\} of Feature Pyramid Network~\cite{lin2017feature}.
Their corresponding spatial resolutions are \{1/16, 1/4, 1\} of the input image size.

\subsection{Loss Function}
The cascade cost volume with $N$ stages produces $N-1$ intermediate outputs and a final prediction.
We apply the supervision to all the outputs and the total loss is defined as:
\begin{equation}
Loss = \sum_{k=1}^{N}\lambda^{k} \cdot L^{k}    
\label{eq_loss}
\end{equation}
where $L^{k} $ refers to the loss at the $k^{th}$ stage and $\lambda^{k}$ refers to its corresponding loss weight.
We adopt the same loss function $L^{k}$ as the baseline networks in our experiments.

\section{Experiments} 
We evaluate the proposed cascade cost volume on multi-view stereo and stereo matching tasks.

\begin{table}[t]
\begin{center}
\footnotesize
\resizebox{\linewidth}{!}{
\begin{tabular}{c|c c c c c }
\toprule[1pt]
\multirow{1}{*}{Methods} & Acc.(mm) & Comp.(mm) &Overall(mm) &GPU Mem(MB) &Run-time(s)   \\ \hline
Camp\cite{campbell2008using_mvsnet3}      
&0.835      &0.554       &0.695   & -    & -\\ 
Furu\cite{furukawa2009accurate_7}          
&0.613      &0.941       &0.777  & -    & -\\ 
Tola\cite{tola2012efficient_mvsnet35}      
&0.342      &1.190       &0.766  & -     & -\\ 
Gipuma\cite{galliani2015massively_mvsnet8} 
&\textbf{0.283}      &0.873       &0.578  & -     & -\\ 
SurfaceNet\cite{ji2017surfacenet_mvsnet14} 
&0.450      &1.040       &0.745 & -     & -\\ 
R-MVSNet\cite{yao2019recurrent}     
&0.383      &0.452       &0.417  &7577     &1.28\\ 
P-MVSNet~\cite{Luo_2019_ICCV}               
&0.406      & 0.434      &0.420  & -     & -\\ 
Point-MVSNet~\cite{chen2019point}           
&0.342       &0.411      &0.376 &8731     &3.35\\ 
MVSNet(D=192)\cite{yao2018mvsnet}                 
&0.456      &0.646       &0.551   &10823     &1.210\\  
MVSNet+Ours                        
&0.325     &0.385      &0.355    &5345     &0.492\\  \hline
Comp. with MVSNet &28.7$\%$ &40.4$\%$  & 35.6$\%$  &50.6$\%$  &59.3$\%$ \\
\bottomrule[1pt]
\end{tabular}}
\end{center}
\vspace{-5mm}
\caption{Multi-view stereo quantitative results of different methods on DTU dataset~\cite{aanaes2016dtu} (lower is better). We conduct this experiment using two resolution settings according to PointMVSNet~\cite{chen2019point} where MVSNet+Ours uses resolution of 1152 $\times$ 864.} 
\vspace{-6mm}
\label{tab:mvs_dtu}
\end{table}

\begin{table*}[t]
\begin{center}
\footnotesize
\resizebox{.9\linewidth}{!}{
\begin{tabular}{c| c c |c c c c c c c c }
\toprule[1pt]
                            &Rank  &Mean   &Family   &Francis &Horse  &Lighthouse  &M60   &Panther   &Playground   &Train    \\  \midrule[0.8pt]

COLMAP~\cite{schonberger2016pixelwise_mvsnet32, schonberger2016structure_p36}                    
                                &54.62  &42.14  &50.41  &22.25  &25.63  &56.43  &44.83  &46.97  &48.53  &42.04    \\
R-MVSNet~\cite{yao2019recurrent}  
                                &40.12  &48.40  &69.96  &46.65  &32.59  &42.95  &51.88  &48.80  &52.00  &42.38   \\
Point-MVSNet~\cite{chen2019point}
                                &38.12	&48.27  &61.79	&41.15  &34.20  &50.79  &51.97  &50.85  &52.38  &43.06   \\
ACMH~\cite{xu2018multi_p42}      &15.00  &54.82  &69.99  &49.45  &45.12  &59.04  &52.64  &52.37  &58.34  &51.61  \\

P-MVSNet~\cite{Luo_2019_ICCV}    &12.25   &55.62	&70.04  &44.64  &40.22  &{\bf 65.20}  &55.08  &{\bf 55.17}  &{\bf 60.37}  &{\bf 54.29}    \\
MVSNet~\cite{yao2018mvsnet}      &52.00  &43.48  &55.99  &28.55  &25.07  &50.79  &53.96  &50.86  &47.90  &34.69   \\  \hline
MVSNet+Ours                    &{\bf 9.50} &{\bf 56.42}  &{\bf 76.36}	&{\bf 58.45}&{\bf 46.20}  &55.53	&{\bf 56.11}  &54.02  &58.17  &46.56 \\
\bottomrule[1pt]
\end{tabular}}
\end{center}
\vspace{-5mm}
\caption{Statistical results on the Tanks and Temples dataset~\cite{knapitsch2017tanks} of state-of-the-art multi-view stereo and our methods.}
\label{tab:mvs_tank}
\end{table*}

\begin{figure*}[t]
\begin{center}
{\includegraphics[width=0.9\linewidth]{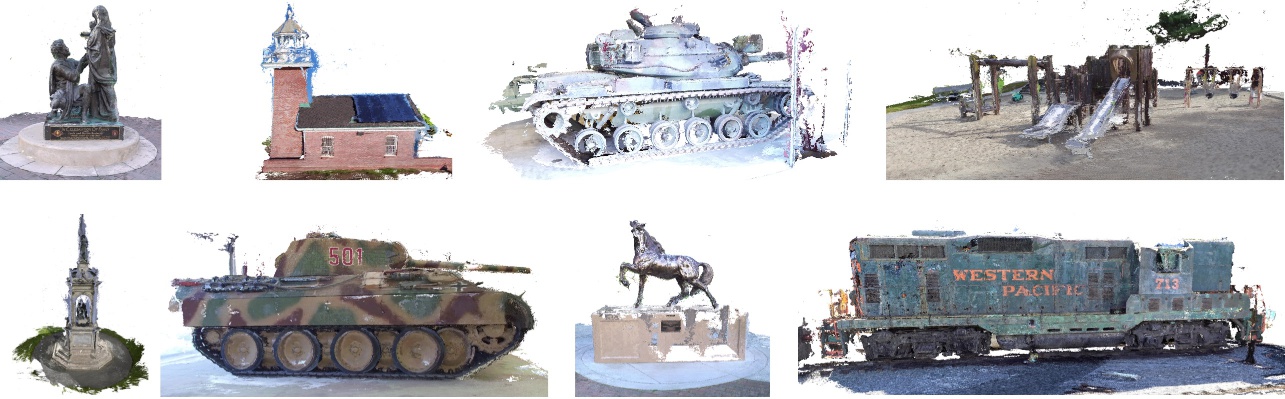}}
\end{center}
\vspace{-5mm}
 \caption{Point cloud results of MVSNet+Ours on the intermediate set of Tanks and Temples dataset~\cite{knapitsch2017tanks}.}
\label{fig:tanks_vis}
\vspace{-3mm}
\end{figure*}

\begin{table}[t!]
\begin{center}
\footnotesize
\resizebox{\linewidth}{!}{
\begin{tabular}{c| c c c c c c}
\toprule[1pt]
Stages           &Resosution     &\textgreater{2mm}(\%) &\textgreater{8mm}(\%)  &Overall (mm)  &GPU Mem. (MB) &Run-time (s)   \\ \hline

1   &1/4 $\times$ 1/4      &0.310   &0.163      &0.602    &2373   &0.081\\
2   &1/2 $\times$ 1/2      &0.208   &0.084      &0.401    &4093   &0.243\\
3   &1                     &0.174   &0.077      &0.355    &5345   &0.492\\
\bottomrule[1pt]
\end{tabular}}
\end{center} 
\vspace{-5mm}
\caption{The statistical results of different stages in cascade cost volume.
The statistics are collected on the DTU evaluation set~\cite{aanaes2016dtu} using MVSNet+Ours.
The run-time is the sum of the current and previous stages. The base of resolution of input images in this experiment is 1152 $\times$ 864.
}
\label{tab:exp_runtime}
\end{table}

\begin{figure}[]
\begin{center}
 {\includegraphics[width=0.24\linewidth]{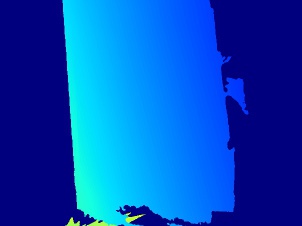}}
 {\includegraphics[width=0.24\linewidth]{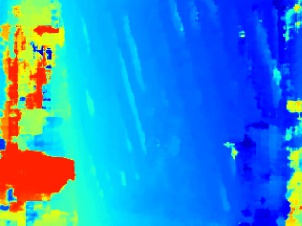}}
{\includegraphics[width=0.24\linewidth]{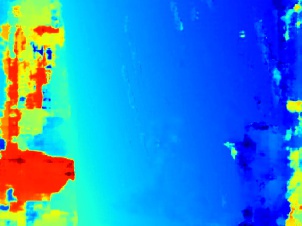}}
{\includegraphics[width=0.24\linewidth]{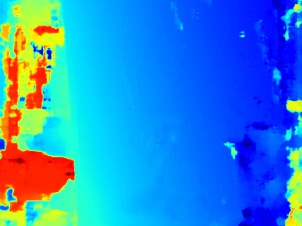}}
 \subfigure[GT$\&$Ref Img]  {\includegraphics[width=0.24\linewidth]{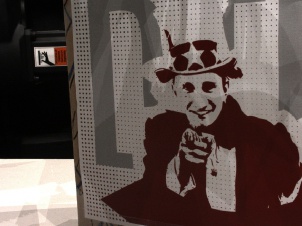}}
  \subfigure[Stage$_1$]  
  {\includegraphics[width=0.24\linewidth]{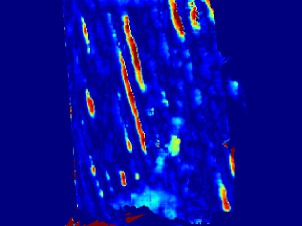}}
 \subfigure[Stage$_2$]  
 {\includegraphics[width=0.24\linewidth]{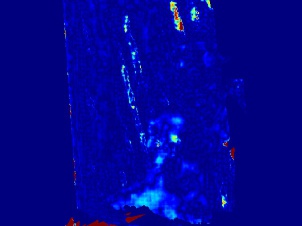}}
 \subfigure[Stage$_3$]  
 {\includegraphics[width=0.24\linewidth]{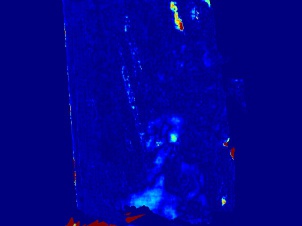}}
\end{center}
\vspace{-8mm}
 \caption{Reconstruction results of each stage. \textbf{Top row}: Ground truth depth map and intermediate reconstructions. \textbf{Bottom row}: Error maps of intermediate reconstructions.}
 \vspace{-6mm}
\label{fig:exp_each_stage}
\end{figure}

\subsection{Multi-view stereo} 
\paragraph{Datasets}
DTU~\cite{aanaes2016dtu} is a large-scale MVS dataset consisting of 124 different scenes scanned in 7 different lighting conditions at 49 or 64 positions. Tanks and Temples dataset~\cite{knapitsch2017tanks} contains realistic scenes with small depth ranges.
More specifically, its intermediate set is consisted of 8 scenes including Family, Francis, Horse, Lighthouse, M60, Panther, Playground, and Train.
Following the work~\cite{yao2019recurrent}, we use DTU training set~\cite{aanaes2016dtu} to train our method,
and test on DTU evaluation set.
To validate the generalization of our approach, we also test it on the intermediate set of Tanks and Temples dataset~\cite{knapitsch2017tanks} using the model trained on DTU dataest without fine-tuning.

\paragraph{Implementation}\vspace{-2mm}
We apply the proposed cascade cost volume to the representative MVSNet~\cite{yao2018mvsnet} and denote the network as \textbf{MVSNet+Ours}.
During training, we set the number of input images to $N$=3 and image resolution to $640\times512$.
After balancing accuracy and efficiency, we adopt a three-stage cascade cost volume.
From the first to the third stage, the number of depth hypothesis is 48, 32 and 8, and the corresponding depth interval is set to 4, 2 and 1  times as the interval of MVSNet~\cite{yao2018mvsnet} respectively.
Accordingly, the spatial resolution of feature maps gradually increases and is set to $1/16$, $1/4$ and $1$ of the original input image size.
We follow the same input view selection and data pre-processing strategies as MVSNet~\cite{yao2018mvsnet} in both training and evaluation.
During training, we use Adam optimizer with $\beta_1 = 0.9$ and $\beta_2 = 0.999$.
The training is done for 16 epochs with an initial learning rate of 0.001, which is downscaled by a factor of 2 after 10, 12, and 14 epochs.
We train our method with 8 Nvidia GTX 1080Ti GPUs with 2 training samples on each GPU. 

For quantitative evaluation on DTU dataset~\cite{aanaes2016dtu}, we calculate the accuracy and the completeness by the MATLAB code provided by DTU dataset~\cite{aanaes2016dtu}. The percentage evaluation is implemented following MVSNet~\cite{yao2018mvsnet}. 
The F-score is used as the evaluation metric for Tanks and Temples dataset \cite{knapitsch2017tanks} to measure the accuracy and completeness of the reconstructed point clouds. 
We use fusibile~\cite{fusible} as our post-processing consisting of three steps: photometric filtering, geometric consistency filtering, and depth fusion.

\begin{figure*}[t!]
\begin{center}
\def\col_k{0.30}
{\includegraphics[width=\col_k\linewidth]{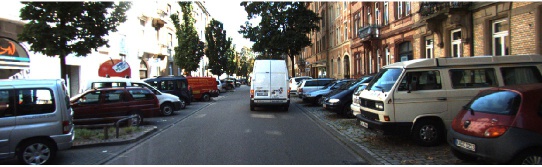}}
{\includegraphics[width=\col_k\linewidth]{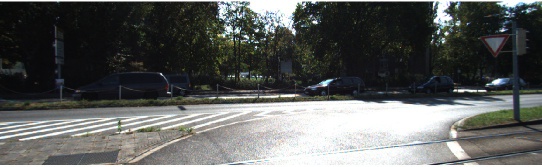}}
{\includegraphics[width=\col_k\linewidth]{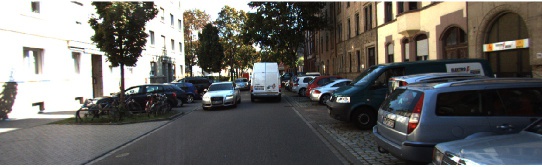}}

{\includegraphics[width=\col_k\linewidth]{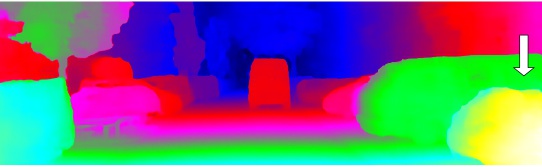}}
{\includegraphics[width=\col_k\linewidth]{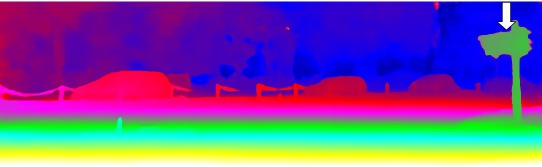}}
{\includegraphics[width=\col_k\linewidth]{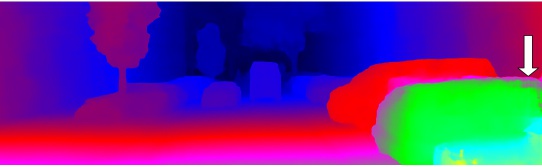}}

{\includegraphics[width=\col_k\linewidth]{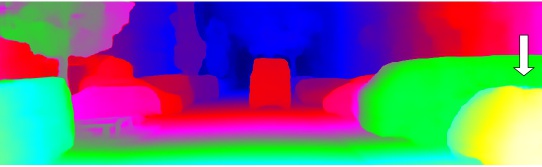}}
{\includegraphics[width=\col_k\linewidth]{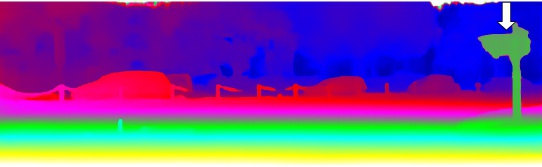}}
{\includegraphics[width=\col_k\linewidth]{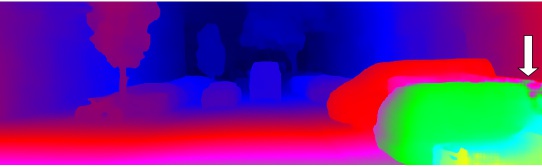}}

{\includegraphics[width=\col_k\linewidth]{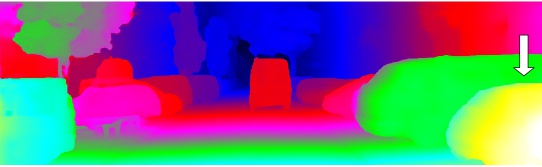}}
{\includegraphics[width=\col_k\linewidth]{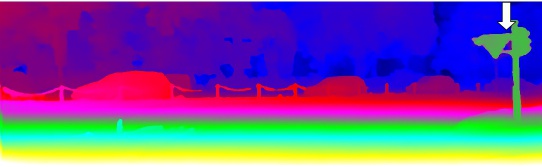}}
{\includegraphics[width=\col_k\linewidth]{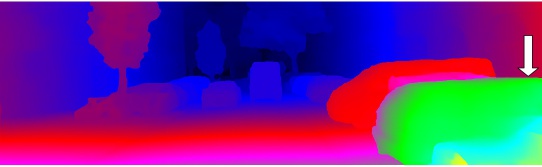}}

\end{center}
\vspace{-5mm}
 \caption{Qualitative results on the test set of KITTI2015~\cite{menze2015object}. \textbf{Top row}: Input images, \textbf{Second row}: Results of PSMNet~\cite{chang2018pyramid}. \textbf{Third row}: Results of GwcNet ~\cite{guo2019group}. \textbf{Bottom row}: Results of GwcNet with cascade cost volume (GwcNet+Ours).}
\label{fig:fig_stereo}
\end{figure*}


\paragraph{Benchmark Performance}\vspace{-1mm}
Quantitative results on DTU evaluation set \cite{aanaes2016dtu} are shown in Table~\ref{tab:mvs_dtu}.
We can see that MVSNet~\cite{yao2018mvsnet} with cascade cost volume outperforms other methods~\cite{chen2019point, Luo_2019_ICCV, yao2018mvsnet, yao2019recurrent} in both completeness and overall quality and rank the $1st$ place on DTU dataset~\cite{aanaes2016dtu}, with the improvement of 35.6\%, and the decrease of memory, run-time reduction of 50.6\% and 59.3\%.
The qualitative results are shown in Figure~\ref{fig:fig_mvs}.
We can see that MVSNet+Ours generates more complete point clouds with finer details. Besides, we demonstrate the generalization ability of our trained model by testing on Tanks and Temples dataset~\cite{knapitsch2017tanks}.
The corresponding quantitative results are reported in Table~\ref{tab:mvs_tank}, and  MVSNet+Ours achieves the state-of-the-art performance among the learning-based multi-view stereo methods.
The qualitative point cloud results of the intermediate set of Tanks and Temples benchmark~\cite{knapitsch2017tanks} are visualized in Figure~\ref{fig:tanks_vis}.
Note that, we get the results of above mentioned methods by running their provided pre-trained model and code except R-MVSNet~\cite{yao2019recurrent} which provides point cloud results with their post-processing method. 

To analyse the accuracy, GPU memory and run-time at each stage, we evaluate the MVSNet+Ours method on the DTU dataset~\cite{aanaes2016dtu}.
We provide comprehensive statistics in Table~\ref{tab:exp_runtime} and visualization results in Figure~\ref{fig:exp_each_stage}.
In a coarse-to-fine manner, the overall quality is improved from 0.602 to 0.355.
Accordingly, the GPU memory increases from 2,373 MB to 4,093 MB and 5,345 MB,
and run-time increases from 0.081 s to 0.243 s and 0.492 s. 


\subsection{Stereo Matching} 
\paragraph{Datasets}
Scene Flow dataset~\cite{mayer2016large} is a large scale-dataset containing 35,454 training and 4,370 testing stereo pairs of size 960 $\times$ 540.
It contains accurate ground truth disparity maps.
We use the Finalpass of the Scene Flow dataset~\cite{mayer2016large} since it contains more motion blur and defocus and is more like a real-world environment.
KITTI 2015~\cite{menze2015object} is a real-world dataset with dynamic street views.
It contains 200 training pairs and 200 testing pairs. Middlebury~\cite{middle} is the publicly available dataset for high-resolution stereo matching contains 60 pairs under imperfect calibration, different exposures, and different lighting conditions.

\paragraph{Implementation}\vspace{-3mm}
In Scene Flow dataset, we extend PSMNet~\cite{chang2018pyramid}, GwcNet~\cite{guo2019group} and GANet11~\cite{zhang2019ga} with our proposed cascade cost volume and denote them as \textbf{PSMNet+Ours}, \textbf{GwcNet+Ours} and \textbf{GANet11+Ours}.
Balancing the trade-off between accuracy and efficiency, a two-stage cascade cost volume is applied, and the number of disparity hypothesis is 12.
The corresponding disparity interval is set to 4 and 1 pixels respectively.
The spatial resolution of feature maps increases from $1/16$ to $1/4$ of the original input image size.
The maximum disparity is set to 192.

In KITTI 2015 benchmark~\cite{menze2015object}, we mainly compare GwcNet~\cite{guo2019group} and GwcNet+Ours.
For a fair comparison, we follow the training details of the original networks.  
The evaluation metric in Scene Flow dataset~\cite{mayer2016large} is end-point-error (EPE), which is the mean absolute disparity error in pixels.
For KITTI 2015~\cite{menze2015object}, the percentage of disparity outliers $D1$ is used to evaluate disparity error larger than \textit{max(3px, 0.05d$^*$)}, where $d^*$ denotes the ground-truth disparity.

\begin{table}[t]
\begin{center}
\footnotesize
\resizebox{0.85\linewidth}{!}{
\begin{tabular}{c|c|c|c|c|c}
\toprule[1pt]
              &\textgreater1px  &\textgreater2px.   &\textgreater3px   &EPE     &Mem.   \\  \hline 
PSMNet~\cite{chang2018pyramid}  
&9.46    &5.19   &3.80   &0.887   &6871     \\   
PSMNet+Ours        
&7.44   &4.61    &3.50   &0.721   &4124    \\  \hline 
GwcNet~\cite{guo2019group}     
&8.03   &4.47    &3.30   &0.765   &7277      \\
GwcNet+Ours     
&7.46   &4.16    &3.04   &0.649   &4585   \\ \hline 
GANet11~\cite{zhang2019ga}        
&  -    &-       &-      &0.95    &6631       \\
GANet11+Ours          
&11.0   &5.97    &4.28   &0.90   &5032 \\ 
\bottomrule[1pt]
\end{tabular}}
\end{center}
\vspace{-5mm}
\caption{Quantitative results of different stereo matching methods with and without cascade cost volume on Scene Flow dataset~\cite{mayer2016large}. Accuracy, GPU memory consumption and run-time are included for comparisons.}
\label{tab:stereo_sceneflow}
\end{table}

\begin{table}[t]
\begin{center}
\footnotesize   
\resizebox{0.85\linewidth}{!}{
\begin{tabular}{c|c c c|c c c}
\toprule[1pt]
\multirow{2}{*}{Methods}   & \multicolumn{3}{c|}{All (\%)}                          & \multicolumn{3}{c}{Noc (\%)} \\ \cline{2-7} 
                                          & D1-bg  & D1-fg  & D1-all     & D1-bg  & D1-fg  & D1-all   \\
\hline
DispNetC~\cite{mayer2016large}            &4.32    &4.41    &4.34        & 4.11   & 3.72   &4.05      \\ 
GC-Net~\cite{kendall2017end}              &2.21    &6.16    &2.87        &2.02    &5.58    &2.61      \\    
CRL~\cite{pang2017cascade}                &2.48    &{\bf 3.59}    &2.67        &2.32    &{\bf 3.12}    &2.45     \\ 
iResNet-i2e2~\cite{liang2018learning}     &2.14    &3.45    &2.36        &1.94    &3.20    &2.15      \\ 
SegStereo~\cite{yang2018segstereo}        &1.88    &4.07    &2.25        &1.76   &3.70     &2.08       \\ 
PSMNet~\cite{chang2018pyramid}            &1.86    &4.62    &2.32        &1.71    &4.31    &2.14      \\ 
GwcNet~\cite{guo2019group}                &1.74    &3.93    &2.11        &1.61    &3.49    &1.92      \\   \hline
GwcNet+Ours                               &{\bf 1.59}    &4.03    &{\bf 2.00}        &{\bf 1.43}    &3.55    &{\bf 1.78}      \\   
\bottomrule[1pt]
\end{tabular}}
\end{center}
\vspace{-5mm}
\caption{Comparison of different stereo matching methods on KITTI2015 benchmark~\cite{menze2015object}.} 
\vspace{-6mm}
\label{tab:stereo_kitti2015}
\end{table}

\paragraph{Benchmark Performance}\vspace{-4mm}
Quantitative results of different stereo methods on Scene Flow dataset~\cite{mayer2016large} is shown in Table~\ref{tab:stereo_sceneflow}.
By applying the cascade 3D cost volume, we boost the accuracy in all the metrics and less memory is required owing to the cascade design with smaller number of disparity hypothesis. 
Our method reduces the end-point-error by 0.166, 0.116 and 0.050 on PSMNet~\cite{chang2018pyramid} (0.887 vs. 0.721), GwcNet~\cite{guo2019group} (0.765 vs. 0.649) and GANet11~\cite{zhang2019ga} (0.950 vs. 0.900) respectively.
The obvious improvement on $\textgreater{1}px$ indicates that small errors are suppressed with the introduction of high-resolution cost volumes. 
In KITTI 2015~\cite{menze2015object}, Table~ \ref{tab:stereo_kitti2015} shows the percentage of disparity outliers $D1$ evaluated for background, foreground, and all pixels.
Compared with the original GwcNet~\cite{guo2019group}, the rank of GwcNet+Ours rises from $29^{th}$ to $17^{th}$ (date: Nov.5, 2019).
Several disparity estimation on KITTI 2015 test set~\cite{menze2015object} is shown in Figure~\ref{fig:fig_stereo}. In Middlebury benchmark, PSMNet+Ours ranks 37th for the avgerr metric(date: Feb.7, 2020).


\begin{table}[t!]
\begin{center}
\footnotesize
\resizebox{\linewidth}{!}{
\begin{tabular}{c|c|c|c c c}
\toprule[1pt]
                                    &Depth Num.  &Depth Interv. &Acc.  &Comp. &Overall\\  
\midrule[0.5pt]
\multirow{1}*{MVSNet}             & 192 & 1                      &0.4560 &0.6460 &0.5510    \\ \midrule[0.5pt]
\multirow{1}*{MVSNet-Cas$_2$}      &96, 96 &2, 1             &{\bf 0.4352} &0.4275 &0.4314     \\ \midrule[0.5pt]
\multirow{1}*{MVSNet-Cas$_3$}      &96, 48, 48 &2, 2, 1       &0.4479 &{\bf 0.4141 }&{\bf 0.4310}    \\    \midrule[0.5pt]
\multirow{1}*{MVSNet-Cas$_4$}      &96, 48, 24, 24 &2, 2, 2, 1&0.4354 &0.4374 &0.4364 \\    \midrule[0.5pt]
\multirow{1}*{MVSNet-Cas$_3$-share}&96, 48, 48 &2, 2, 1       &0.4741 &0.4282  &0.4512    \\   
\bottomrule[1.pt]
\end{tabular}}
\end{center} 
\vspace{-5mm}
\caption{Comparisons between MVSNet~\cite{yao2018mvsnet} and MVSNet using our cascade cost volume with different setting of depth hypothesis numbers and depth intervals. 
The statistics are collected on DTU dataset~\cite{aanaes2016dtu}.}
\vspace{-1mm}
\label{tab:exp_cas_num}
\end{table}

\begin{table}[t]
\begin{center}
\footnotesize
\resizebox{1.0\linewidth}{!}{
\begin{tabular}{c | c c c | c c c}
\toprule[1pt]
Methods                &cascade? &upsample?  &feature pyramid? &Acc. (mm)  &Comp. (mm) &Overall (mm)  \\ \hline
MVSNet                 &$\times$ &$\times$   &$\times$    & 0.456    &0.646  &0.551   \\   
MVSNet-Cas$_3$         &$\checkmark$ &$\times$ &$\times$  &0.450     &0.455  &0.453    \\
MVSNet-Cas$_3$-Ups     &$\checkmark$ &$\checkmark$  &$\times$     &0.419     &0.338  &0.379   \\ 
MVSNet+Ours            &$\checkmark$  &$\times$ &$\checkmark$   &0.325     &0.385  &0.355     \\
\bottomrule[1pt]
\end{tabular}
}
\end{center} 
\vspace{-5mm}
\caption{The quantitative comparison between MVSNet and MVSNet with different settings of the cascade cost volumes. Specifically, "cascade" denotes that the original cost volume is divided into three cascade cost volumes, "upsample" denotes cost volumes with increased spatial resolutions by bilinear upsampling corresponding feature maps, and “feature pyramid” denotes cost volumes with higher spatial resolutions built on pyramid feature maps. The statistics are evaluated on the DTU dataset.}
\vspace{-4mm}
\label{tab:exp_cas_spatial}
\end{table}


\subsection{Ablation Study} 
Extensive ablation studies are performed to validate the improved accuracy and efficiency of our approach.
All results are obtained by the three-stage model on DTU validation set~\cite{aanaes2016dtu} unless otherwise stated.

\paragraph{Cascade Stage Number}\vspace{-4mm}
The quantitative results with different stage numbers are summarized in Table~\ref{tab:exp_cas_num}. In our implementation, we use MVSNet~\cite{yao2018mvsnet} with 192 depth hypothesis as the baseline model, and replace its cost volume with our cascade design which is also consisted of 192 depth hypothesis.
Note that the spatial resolution of different stages are the same as that of the original MVSNet~\cite{yao2018mvsnet}.
This extended MVSNet is denoted as \textbf{MVSNet-Cas$_i$} where $i$ indicates the total stage number.
We find that as the number of stages increases, the overall quality first remarkably increases and then stabilizes.

\paragraph{Spatial Resolution}\vspace{-4mm}
Then, we study how the spatial resolution of a cost volume $W\times H$ affects the reconstruction performance.
Here, we compare MVSNet-Cas$_3$,
which contains 3 stages and all the stages share the same spatial resolution,
and MVSNet-Cas$_3$-Ups where the spatial resolution increases from $1/16$ to $1$ of the original image size and bilinear interpolation is used to upsample feature maps.
As shown in Table~\ref{tab:exp_cas_spatial}, the overall quality of MVSNet+Ours is obviously superior to those of MVSNet-Cas$_3$ (0.453 vs. 0.355).
Accordingly, a higher spatial resolution also leads to increased GPU memory (2373 vs. 5345 MB) and run-time (0.322 vs. 0.492 seconds).

\paragraph{Feature Pyramid}\vspace{-4mm}
As shown in Table~\ref{tab:exp_cas_spatial}, the cost volume constructed from Feature Pyramid Network~\cite{lin2017feature} denoted by MVSNet+Ours can slightly improve the overall quality from 0.379 to 0.355.
The GPU memory (6227 vs. 5345 MB) and run-time (0.676 vs. 0.492 seconds) are also decreased.
Compared with the improvement between MVSNet-Cas$_3$ and MVSNet-Cas$_3$-Ups, the increased spatial resolution is still more critical to the improvement of reconstruction accuracy.

\paragraph{Parameter Sharing in Cost Volume Regularization}\vspace{-4mm}
We also analyze the effect of weight sharing in 3D cost volume regularization across all the stages.
As is shown in Table~\ref{tab:exp_cas_num}, the shared parameters cascade cost volume denoted by MVSNet-Cas$_3$-share achieves worse performance than MVSNet-Cas$_3$.
It indicates that separate parameter learning of the cascade cost volumes at different stages further improves the accuracy.


\subsection{Run-time and GPU Memory} 
Table~\ref{tab:mvs_dtu} shows the comparison of GPU memory and run-time between MVSNet~\cite{yao2018mvsnet} with and without cascade cost volume.
Given the remarkable accuracy improvement, the GPU memory decreases from 10,823 to 5,345 MB, and the run-time drops from 1.210 to 0.492 seconds.
In Table~\ref{tab:stereo_sceneflow}, we compare the GPU memory between PSMNet~\cite{chang2018pyramid}, GwcNet~\cite{guo2019group} and GANet11~\cite{zhang2019ga} with and without the proposed cascade cost volume.
The GPU memory of PSMNet~\cite{chang2018pyramid}, GwcNet~\cite{guo2019group} and GANet11~\cite{zhang2019ga} decreases by $39.97\%$, $36.99\%$ and $24.11\%$ respectively.




\section{Conclusion} 
In this paper, we present a both GPU memory and computationally efficient cascade cost volume formulation for high-resolution multi-view stereo and stereo matching.
First, we decompose the single cost volume into a cascade formulation of multiple stages.
Then, we can narrow the depth (or dispartiy) range of each stage and reduce the total number of hypothesis planes by utilizing the depth (or disparity) map from the previous stage.
Next, we use the cost volumes of higher spatial resolution to generate the outputs with finer details.
The proposed cost volume is complementary to existing 3D cost-volume-based multi-view stereo and stereo matching approaches. 


{\small
\bibliographystyle{ieee_fullname}
\bibliography{egpaper_for_review}
}

\clearpage 
\begin{figure*}[ht!]
\begin{center}
\subfigure[Distribution of absolute errors at the 1$_{st}$ stage]{
{\includegraphics[width=0.48\linewidth]{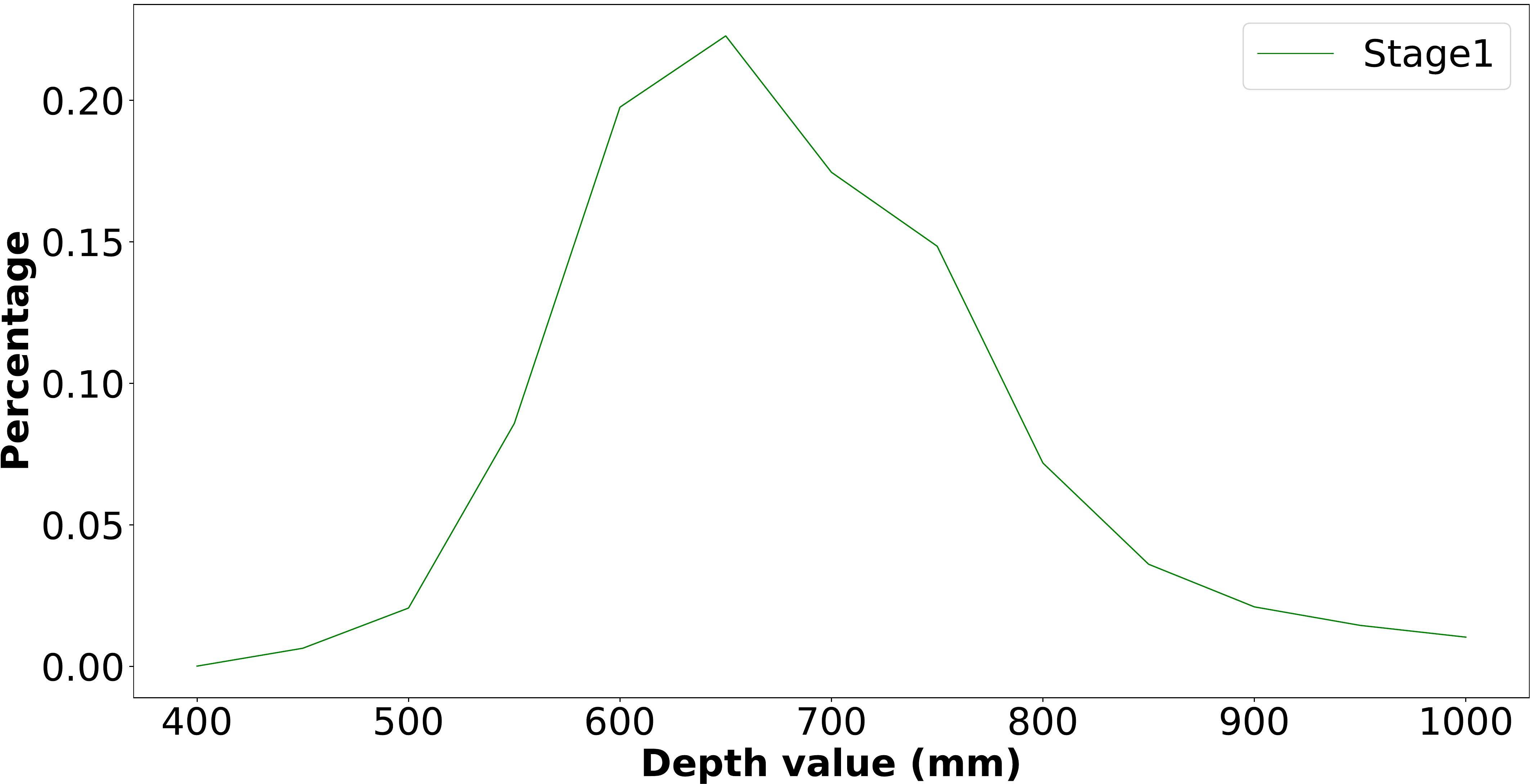}}
}
\subfigure[Distribution of absolute errors at the 2$_{nd}$ and 3$_{rd}$ stages]{
{\includegraphics[width=0.48\linewidth]{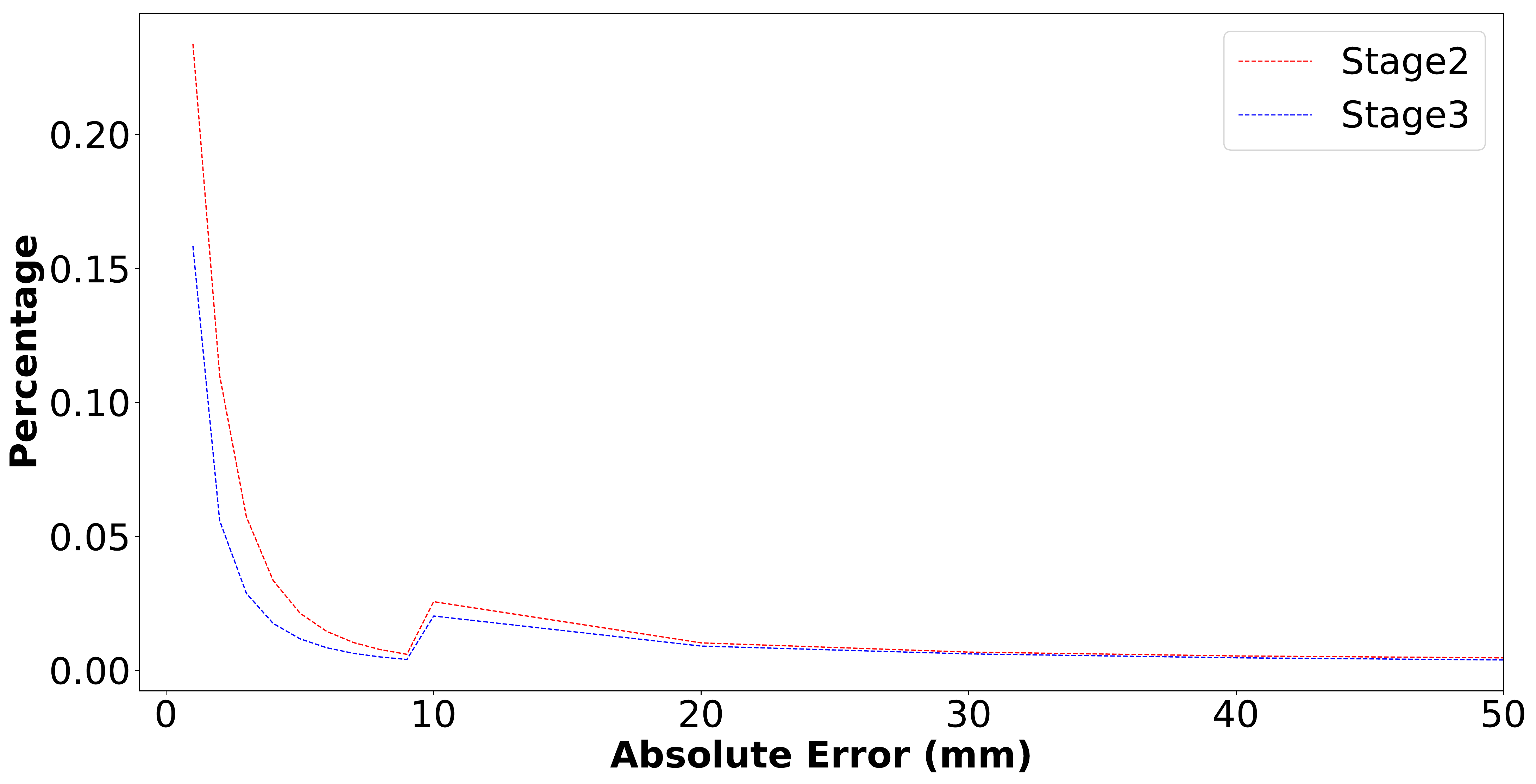}}
}
\end{center}
\vspace{-4mm}
\caption{Distribution of absolute errors at different stages.
We assume that the absolute errors at the 1$_{st}$ stage are the ground-truth depth since there is no predicted depth at this stage.
The statistical results are calculated on DTU evaluation dataset~\cite{aanaes2016dtu} using MVSNet+Ours with a three-stage cost volume.}
\label{fig:search space}
\end{figure*}

\section{Appendix}
\subsection{Discussion}

\paragraph{Why Hypothesis Range is Remarkably Decreased?}\label{Why_Hypothesis_Range}
In Figure~\ref{fig:search space}, we provide the statistics of the absolute depth errors which measure the distance between the predicted depth and its ground truth.
Since there is no depth prediction at the first stage, we regard the ground-truth depth as the absolute depth errors of the first stage.
As shown in Figure~\ref{fig:search space}(a), the entire depth range of the first stage is approximately 500mm while the entire depth range at the second and the third stage shown in Figure~\ref{fig:search space}(b) is narrowed to about 50mm which is reduced by 90\% compared with the first stage. Accordingly, we can significantly reduce the hypothesis range at the second and third stages.

\vspace{-3mm}

\begin{figure*}[t!]
\begin{center}
\def\col_k{0.44}
\subfigure[Results on DTU dataset~\cite{aanaes2016dtu}]  
{\includegraphics[width=0.48\linewidth]{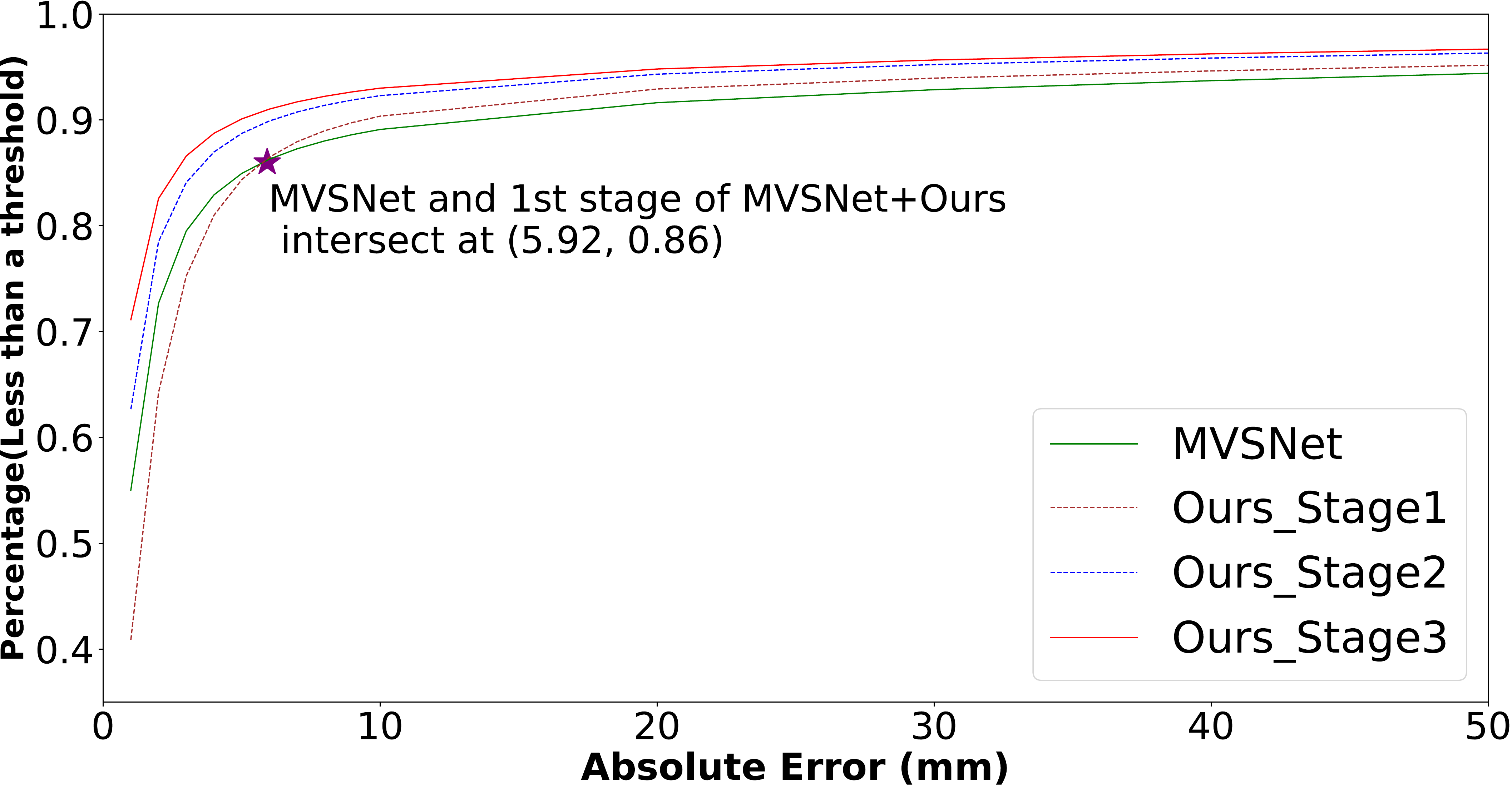}}
\subfigure[Results on Scene Flow dataset~\cite{aanaes2016dtu}]  
{\includegraphics[width=0.48\linewidth]{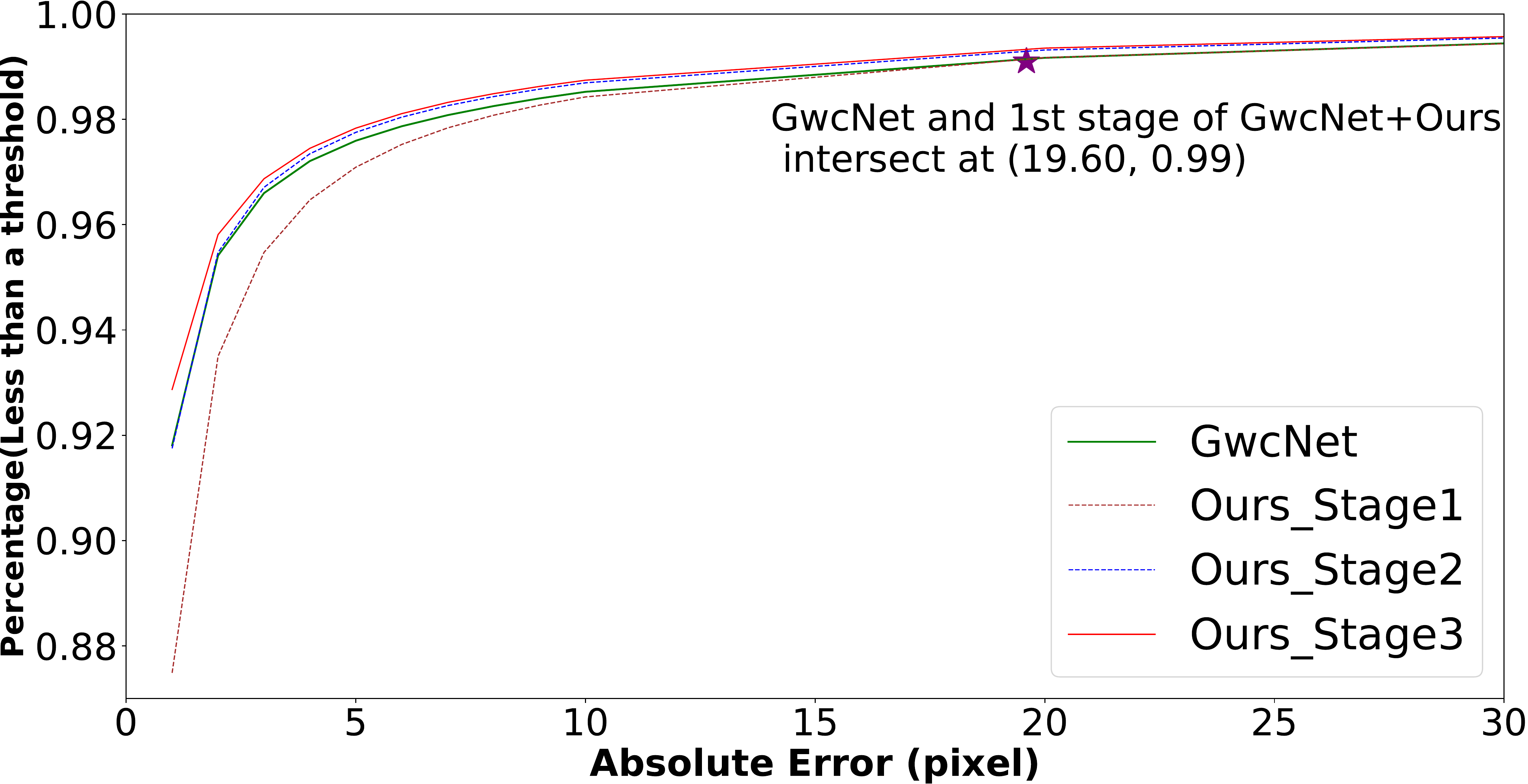}}
\end{center}
\vspace{-4mm}
 \caption{The percentage of the absolute errors between the prediction and the ground-truth less than a certain threshold.
We demonstrate the results of MVSNet~\cite{yao2018mvsnet}, GwcNet~\cite{guo2019group}, and certain networks with cascade cost volume at different stages.}
\label{fig:hypothesisrange}
\vspace{-4mm}
\end{figure*}

\paragraph{How to Set Hypothesis Range at Different Stages?}\label{effect_of_hypothesis}
In Figure~\ref{fig:hypothesisrange}, we calculate the percentage of the absolute errors less than a certain threshold (noted as inlier percentage). 
Hypothesis range should cover most erroneously predicted depth (or disparity) and correct them. 
As shown in Figure~\ref{fig:hypothesisrange}(a), the inlier percentage of MVSNet~\cite{yao2018mvsnet} and that of MVSNet+Ours at the first stage intersects at 5.92mm and 86\%. 
That means if we set a hypothesis range larger than 5.92mm, we can cover more possibly correct predictions than MVSNet, since our cascade cost volume is able to correct the erroneous prediction at the later stages.
On multi-view stereo data-sets, we set the hypotheis range as $32\times 2\times 2.5 = 160$mm which still has a large margin to be reduced.

Similarly in stereo matching, as shown in Figure~\ref{fig:hypothesisrange}(b), the disparity hypothesis range is set as $12\times 2 =24$ pixel  (the intersection is 19.60 pixel), which covers the range of more erroneous predictions at the first stage compared with the original single cost volume approach.

\vspace{-2mm}

\paragraph{Why Cascade Cost Volume is Memory Efficient?}

In multi-view stereo, the hypothesis range is able to remarkably decrease since the entire depth range is narrowed by nearly 90\% (500mm vs. 50mm) since the first stage. Therefore, we can use less hypothesis planes for cost volumes in later stage. In MVSNet+Ours, we set the number of hypothesis planes in first stage as 48 whereas MVSNet~\cite{yao2018mvsnet} has 192 planes, leading to the GPU memory decrease from 10,823MB to 2,373MB. In order to improve the accuracy in subsequent stages, we increase the spatial resolution and the GPU memory increases from 2,373 MB to 4,093 MB and 5,345 MB. Although we increase the spatial resolution, the total GPU memory is deceased about 50.6\% compared MVSNet and run-time is about 2 times faster shown in Figure 1 in the main paper. Similarly, in stereo matching we also decrease the GPU memory from 3,827MB to 2,699MB using our two stage cost volume.

Moreover, we can balance between the time (or memory) efficiency and accuracy by adopting different cascade numbers, hypothesis range and spatial resolutions. 

\subsection{Multi-view Stereo}
In this section, we demonstrate more multi-view stereo experimental results.
As shown in Figure~\ref{fig:dtu_vis}, we visualize the reconstructed point cloud of MVSNet+Ours on DTU dataset~\cite{aanaes2016dtu}.

\subsection{Stereo Matching}
\paragraph{Qualitative Results on Scene Flow Dataset}
In this section, we show several reconstruction results of PSMNet~\cite{chang2018pyramid}, GwcNet~\cite{guo2019group}, GANet11~\cite{zhang2019ga}  and the extended model \textbf{PSMNet+Ours}, \textbf{GwcNet+Ours}, \textbf{GANet11+Ours} on Scene Flow dataset~\cite{mayer2016large}.  As is shown in Figure~\ref{fig:sceneflow}, the visual quality is improved with the replacement of our cascade cost volume.

\begin{figure*}[t!]
\begin{center}
\def\col_k{0.20}

{\includegraphics[width=\col_k\linewidth]{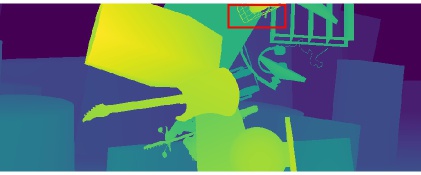}}
{\includegraphics[width=\col_k\linewidth]{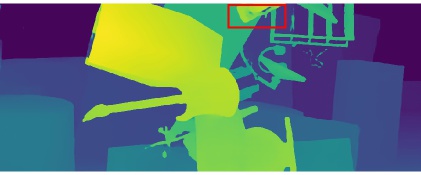}}
{\includegraphics[width=\col_k\linewidth]{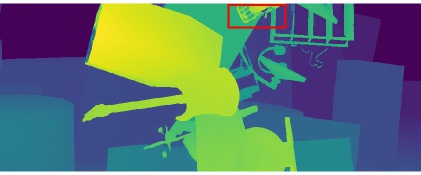}}
{\includegraphics[width=\col_k\linewidth]{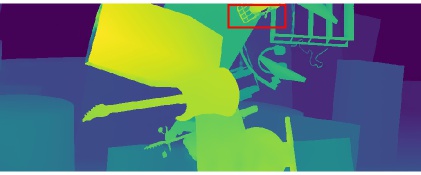}}

\subfigure[Ref Img \& GT]  
{\includegraphics[width=\col_k\linewidth]{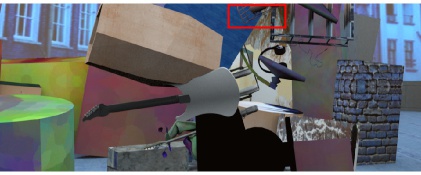}}
\subfigure[Stage1]
{\includegraphics[width=\col_k\linewidth]{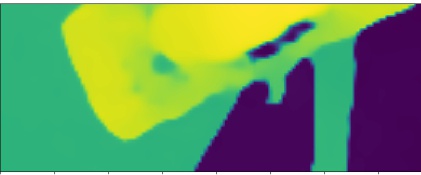}}
\subfigure[Stage2]
{\includegraphics[width=\col_k\linewidth]{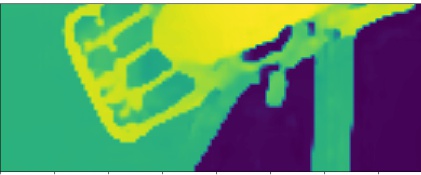}}
\subfigure[Stage3]
{\includegraphics[width=\col_k\linewidth]{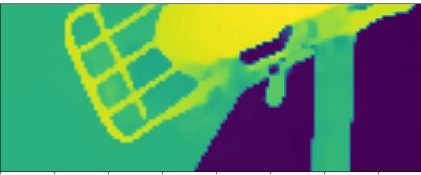}}

{\includegraphics[width=\col_k\linewidth]{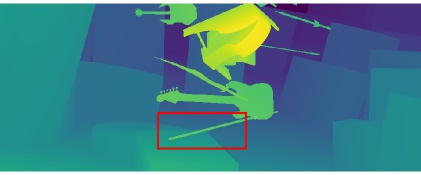}}
{\includegraphics[width=\col_k\linewidth]{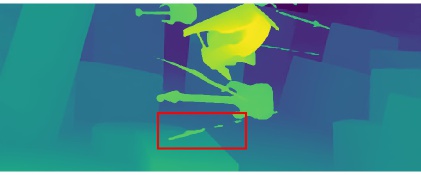}}
{\includegraphics[width=\col_k\linewidth]{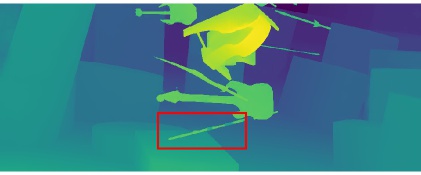}}
{\includegraphics[width=\col_k\linewidth]{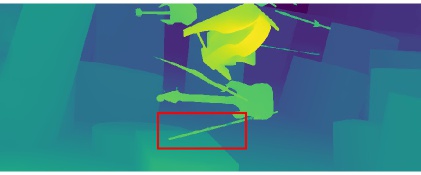}}

\subfigure[Ref Img \& GT]  
{\includegraphics[width=\col_k\linewidth]{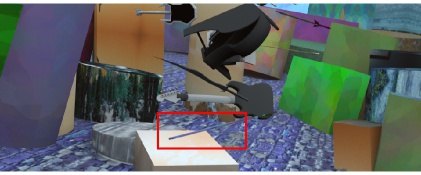}}
\subfigure[Stage1]
{\includegraphics[width=\col_k\linewidth]{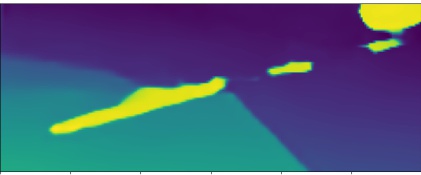}}
\subfigure[Stage2]
{\includegraphics[width=\col_k\linewidth]{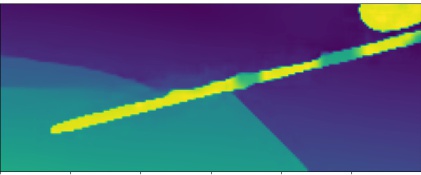}}
\subfigure[Stage3]
{\includegraphics[width=\col_k\linewidth]{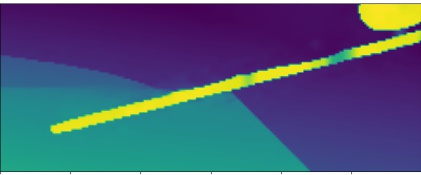}}

{\includegraphics[width=\col_k\linewidth]{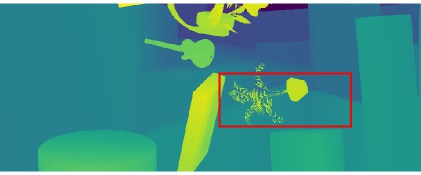}}
{\includegraphics[width=\col_k\linewidth]{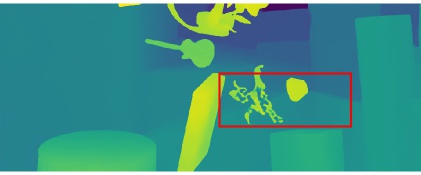}}
{\includegraphics[width=\col_k\linewidth]{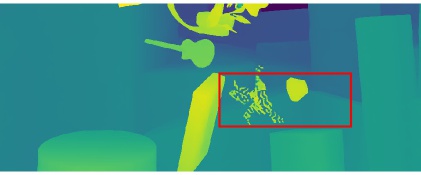}}
{\includegraphics[width=\col_k\linewidth]{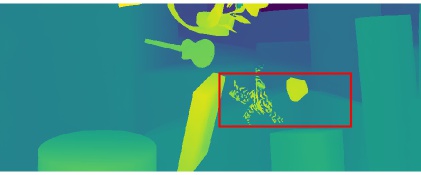}}

\subfigure[Ref Img \& GT]  
{\includegraphics[width=\col_k\linewidth]{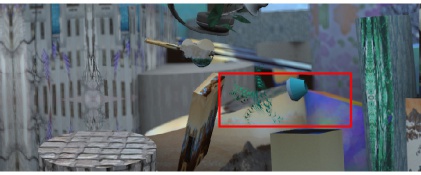}}
\subfigure[Stage1]
{\includegraphics[width=\col_k\linewidth]{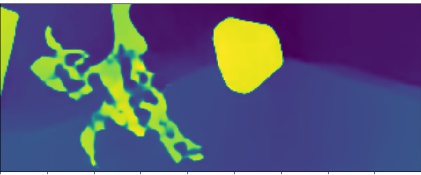}}
\subfigure[Stage2]
{\includegraphics[width=\col_k\linewidth]{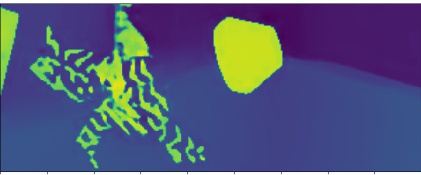}}
\subfigure[Stage3]
{\includegraphics[width=\col_k\linewidth]{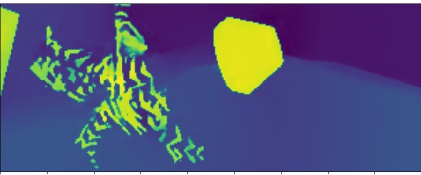}}

{\includegraphics[width=\col_k\linewidth]{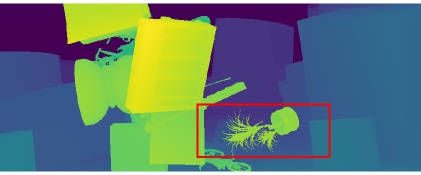}}
{\includegraphics[width=\col_k\linewidth]{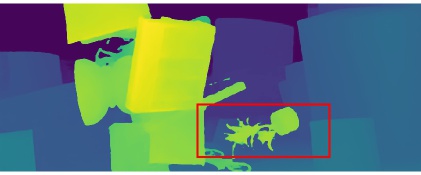}}
{\includegraphics[width=\col_k\linewidth]{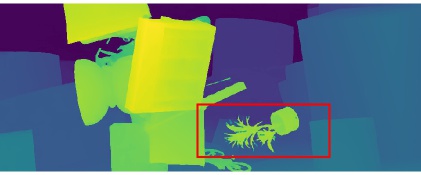}}
{\includegraphics[width=\col_k\linewidth]{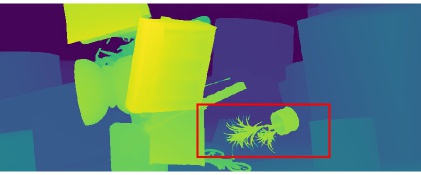}}

\subfigure[Ref Img \& GT]  
{\includegraphics[width=\col_k\linewidth]{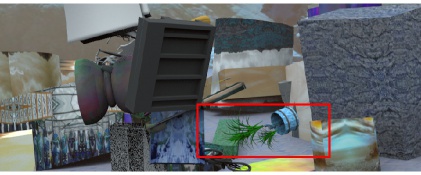}}
\subfigure[Stage1]
{\includegraphics[width=\col_k\linewidth]{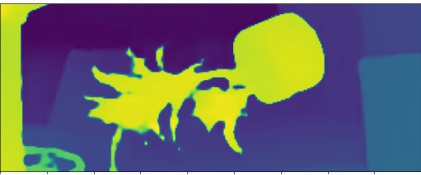}}
\subfigure[Stage2]
{\includegraphics[width=\col_k\linewidth]{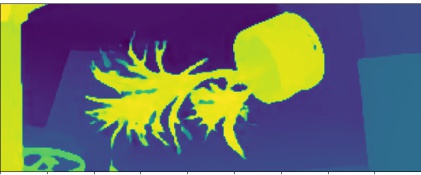}}
\subfigure[Stage3]
{\includegraphics[width=\col_k\linewidth]{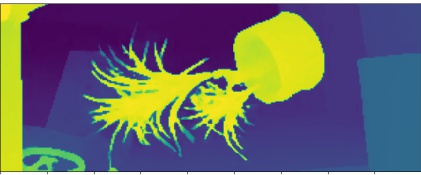}}

\end{center}
 \caption{Reconstruction results of each intermediate stage of GwcNet+Ours on Scene Flow test dataset~\cite{mayer2016large}. From left to right: reference image and ground truth, the predicted disparity of stage1, stage2 and stage3.
 The zoomed areas of intermediate reconstructions is shown below its intermediate reconstructions}
\label{fig:each_stage_sceneflow}
\end{figure*}

\begin{figure*}[t!]
\begin{center}
\def\col_k{0.25}

\subfigure[Ref Img]  
{\includegraphics[width=\col_k\linewidth]{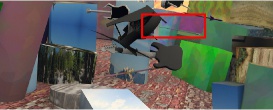}}
\subfigure[PSMNet~\cite{chang2018pyramid}]
{\includegraphics[width=\col_k\linewidth]{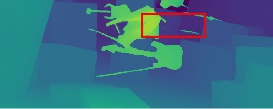}}
\subfigure[Error map of PSMNet~\cite{chang2018pyramid}]
{\includegraphics[width=\col_k\linewidth]{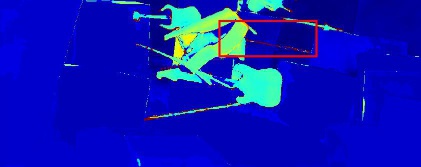}}
\subfigure[GT]  
{\includegraphics[width=\col_k\linewidth]{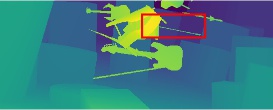}}
\subfigure[PSMNet+Ours]
{\includegraphics[width=\col_k\linewidth]{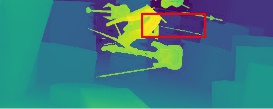}}
\subfigure[Error map of PSMNet+Ours]
{\includegraphics[width=\col_k\linewidth]{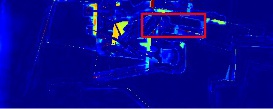}}

\subfigure[Ref Img]  
{\includegraphics[width=\col_k\linewidth]{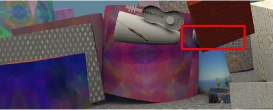}}
\subfigure[GANet11~\cite{zhang2019ga}]
{\includegraphics[width=\col_k\linewidth]{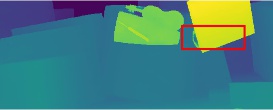}}
\subfigure[Error map of GANet11~\cite{zhang2019ga}]
{\includegraphics[width=\col_k\linewidth]{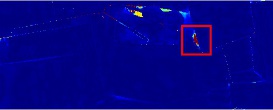}}
\subfigure[GT]  
{\includegraphics[width=\col_k\linewidth]{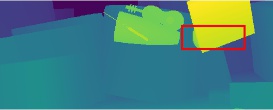}}
\subfigure[GANet11+Ours]  
{\includegraphics[width=\col_k\linewidth]{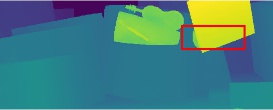}}
\subfigure[Error map of GANet11+Ours]  
{\includegraphics[width=\col_k\linewidth]{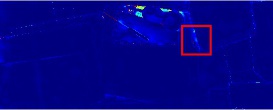}}


\subfigure[Ref Img]  
{\includegraphics[width=\col_k\linewidth]{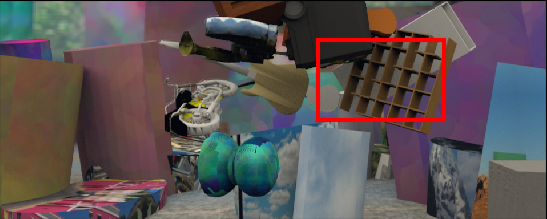}}
\subfigure[GwcNet~\cite{guo2019group}]
{\includegraphics[width=\col_k\linewidth]{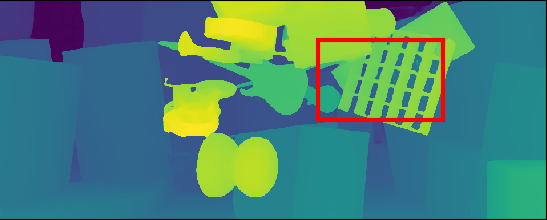}}
\subfigure[Error map of GwcNet~\cite{guo2019group}]
{\includegraphics[width=\col_k\linewidth]{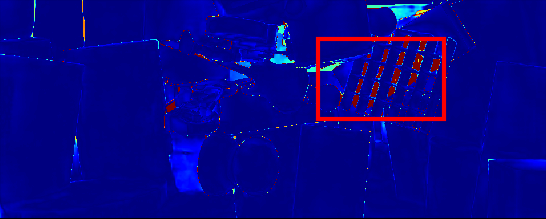}}
\subfigure[GT]  
{\includegraphics[width=\col_k\linewidth]{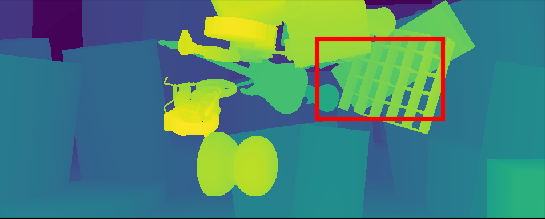}}
\subfigure[GwcNet+Ours]  
{\includegraphics[width=\col_k\linewidth]{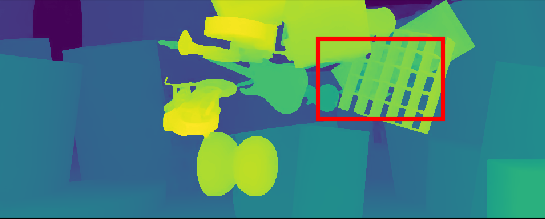}}
\subfigure[Error map of GwcNet+Ours]  
{\includegraphics[width=\col_k\linewidth]{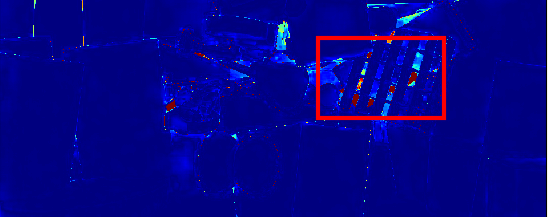}}

\end{center}
\vspace{-8mm}
 \caption{Qualitative results on the test set of Scene Flow dataset~\cite{mayer2016large}. We show the results of several representative stereo CNNs and the extended models with the proposed cascade cost volume.}
\label{fig:sceneflow}
\vspace{-4mm}
\end{figure*}

\vspace{-2mm}
\paragraph{Cascade Stage Number in Stereo Matching}
In this experiment, we replace the cost volume in GwcNet~\cite{guo2019group} with our proposed cascade cost volume, namely GwcNet+Ours. Note that, the experiment setting in GwcNet~\cite{guo2019group} is 64 channel concatenation volume, the spatial resolution of different stages are the same as that of the original GwcNet. The extended model with total i$_{th}$ stages is denoted as \textbf{GwcNet-Cas$_i$}. As is shown in Table~\ref{tab:stereo_cas_num}, the accuracy of the extended model increases with stage increases. We can notice the details get cleaner as the stage increases in Figure~\ref{fig:each_stage_sceneflow}.

\begin{table}[t!]
\begin{center}
\footnotesize
\resizebox{\linewidth}{!}{
\begin{tabular}{c|c|c|c| c cc}
\toprule[1pt]
                                    &Disparity Num.  &Disparity Interv. &EPE(pixel)  &D1(\%)\\  \midrule[0.5pt]
\multirow{1}*{GwcNet}              &48 & 1                     & 0.833  &0.294 \\ \midrule[0.5pt]
\multirow{1}*{GwcNet-Cas$_2$}      &24, 24 &2, 1               & 0.764  &0.283\\ \midrule[0.5pt]
\multirow{1}*{GwcNet-Cas$_3$}      &24, 12, 12 &2, 2, 1        & 0.737  &0.274  \\    \midrule[0.5pt]
\multirow{1}*{GwcNet-Cas$_4$}      &24, 12, 6, 6 &2, 2, 2, 1   & 0.703  &0.264\\    
\bottomrule[1.pt]
\end{tabular}}
\end{center} 
\vspace{-4mm}
\caption{Comparisons between GwcNet~\cite{guo2019group} and GwcNet using our cascade cost volume with different setting of the numbers of hypothesis planes and depth intervals. The statistics are collected on the test set of Scene Flow dataset~\cite{mayer2016large}}
\label{tab:stereo_cas_num}
\vspace{-4mm}
\end{table}

\paragraph{Spatial Resolution in Stereo Matching}

\vspace{-5mm}
We study how the spatial resolution of a cost volume affects the reconstruction accuracy and GPU memory in stereo matching. Similar to the experiment in multi-view stereo, we formulate a three-stage cost volume based on GwcNet with the spatial resolution gradually increases from 1$/$4 $\times$ 1$/$4 to 1 of the original input image size. In a coarse-to fine manner, the end-point-error is improved from 0.972 to 0.619. Accordingly, the GPU memory increases from 1,545MB to 3,429MB.

\begin{table*}[t!]
\begin{center}
\footnotesize
\resizebox{0.65\linewidth}{!}{
\begin{tabular}{c| c c c c c c c}
\toprule[1pt]
Stage          &Resosution     &\textgreater{1}(\%)  &\textgreater{2}(\%)   &\textgreater{3}(\%)   &EPE(pixel)  &D1(\%) &GPU Mem. (MB)   \\ \hline

1                               &1/4 $\times$ 1/4      &12.4 &6.43   &4.46    &0.972     &3.65  &1545    \\
2                               &1/2 $\times$ 1/2      &8.17 &4.45   &3.22     &0.680     &2.62  &2699    \\
3                               &1        &7.06  &4.12   &3.06     &0.619     &2.49    &3429    \\ \hline
GwcNet\cite{guo2019group}     &1/4 $\times$ 1/4    &8.41 &4.63 &3.41 &0.808  &2.84  &3827 \\
\bottomrule[1pt]
\end{tabular}}
\end{center} 
\vspace{-4mm}
\caption{The statistical results of different stages in cascade cost volume.
The statistics are collected on the Scene Flow evaluation set~\cite{mayer2016large} using GwcNet+Ours.
The run-time is the sum of the current and previous stages and the original input size is 960$\times$512.
}
\vspace{-4mm}
\label{tab:exp_runtime}
\end{table*}

\begin{figure*}[ht!]
    \centering
    \def\col_k{0.25}
    \subfigure[Ref Img]  
    {\includegraphics[width=\col_k\linewidth]{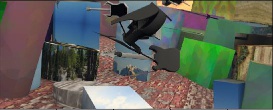}}
    \vspace{-3mm}
    \subfigure[GwcNet~\cite{guo2019group}]
    {\includegraphics[width=\col_k\linewidth]{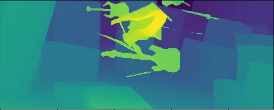}}
    \subfigure[Error map of GwcNet~\cite{guo2019group}]
    {\includegraphics[width=\col_k\linewidth]{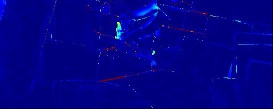}}
    \subfigure[GT]  
    {\includegraphics[width=\col_k\linewidth]{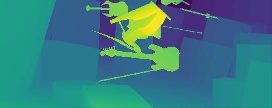}}
    \subfigure[GwcNet+Ours]  
    {\includegraphics[width=\col_k\linewidth]{./imgs/supp/sceneflow_compare/num23_gwcnet.jpg}}
    \subfigure[Error map of GwcNet+Ours]  
    {\includegraphics[width=\col_k\linewidth]{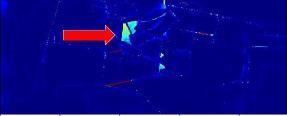}}
    
    \vspace{-5mm}
    \caption{A failed case of GwcNet+Ours on the test set of Scene Flow dataset~\cite{mayer2016large}. \textbf{Top row:} Reference image, the prediction of GwcNet~\cite{guo2019group} and the error map of GwcNet. \textbf{Bottom row:} Ground Truth, the prediction of GwcNet+Ours and the error map of GwcNet+Ours. The red arrow points out the wrong prediction region.}
    \label{fig:failed_case}
\end{figure*}

\begin{figure*}[ht!]
    \centering
    \includegraphics[width=0.85\linewidth]{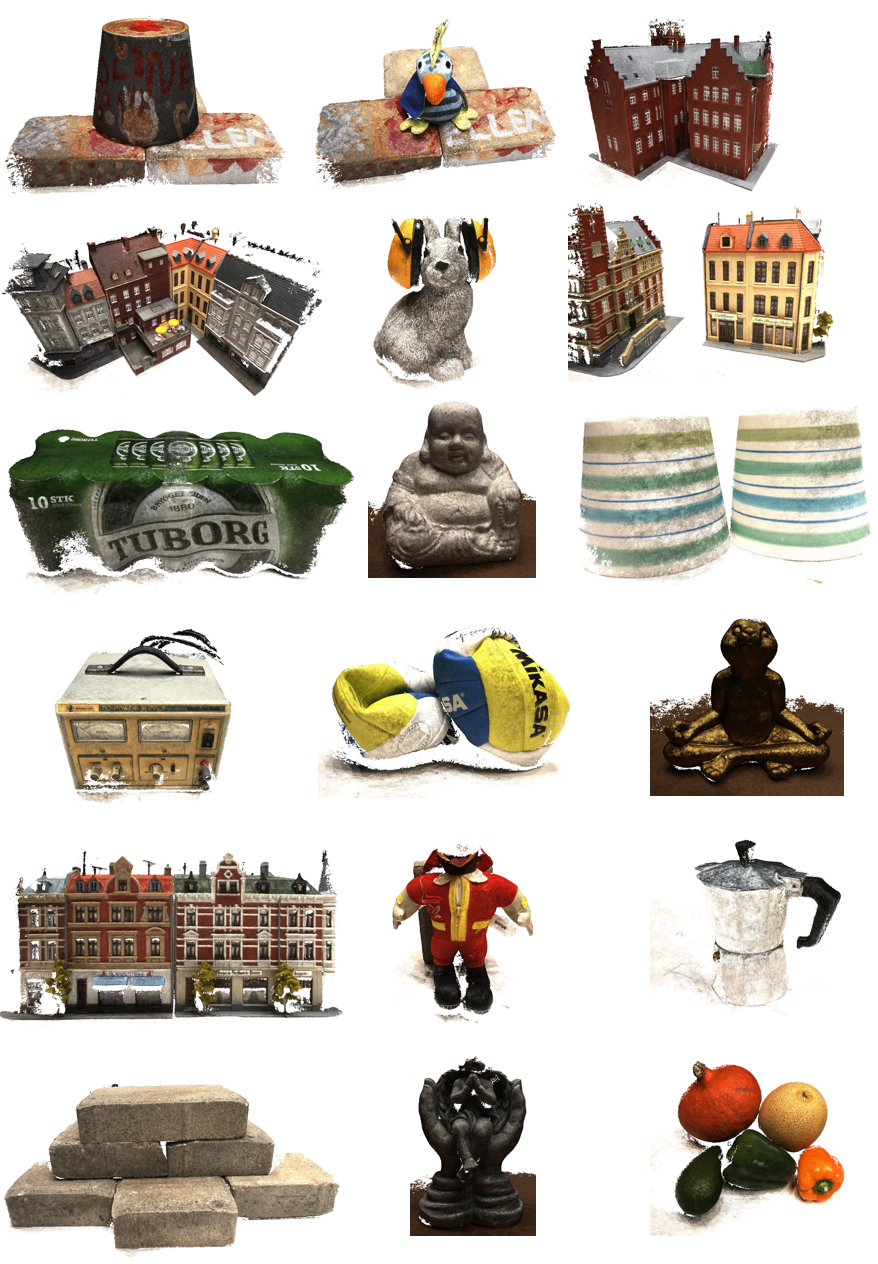}
    \caption{Point cloud results of MVSNet+Ours on DTU evaluation dataset~\cite{aanaes2016dtu}}
    \label{fig:dtu_vis}
\end{figure*}

\subsection{Limitations and Future Works}
The proposed cascade cost volume formulation benefits from decomposing the single cost volume into a cascade formulation of multiple stages. We have analyzed the effect of hypothesis range setting in Section \ref{effect_of_hypothesis}. Although the cascade formulation is complementary to existing 3D cost-volume-based multi-view stereo and stereo matching approaches, some limitations still exist.
As shown in Figure~\ref{fig:failed_case}, GwcNet+Ours generates a biased result since the earlier stages output erroneous disparity and the hypothesis range in the next stage is not able to cover its corresponding ground truth value. Note that this case happens with little probability since the cascade cost volume formulation could correct almost erroneous predictions according to the analysis in Section~\ref{effect_of_hypothesis} and the overall performance is also better than single cost volume models. 

Currently, the hypothesis range of each pixel is identical. The future works include determine the hypothesis range for each region by incorporating semantic information but probably need a more flexible cost volume formulation. 

\end{document}